\documentclass[10pt,twocolumn,letterpaper]{article}

\usepackage{wacv}
\usepackage{times}
\usepackage{epsfig}
\usepackage{graphicx}
\usepackage{amsmath}
\usepackage{amssymb}

\usepackage{microtype}
\usepackage{graphicx}
\usepackage{subfigure}
\usepackage{booktabs} 
\usepackage{soul}

\usepackage{graphicx}
\usepackage{amsmath}
\usepackage{amssymb}
\usepackage{enumitem}
\usepackage{tabularx}
\usepackage{booktabs}
\usepackage[ruled]{algorithm2e}
\usepackage{subfigure}
\usepackage{algorithmic}
\usepackage{amsfonts}       
\usepackage{nicefrac}       
\usepackage{microtype}      
\usepackage[table]{xcolor}
\usepackage{multirow}
\usepackage{comment}
\usepackage[accsupp]{axessibility}

\def\wacvPaperID{1268} 

\wacvfinalcopy 

\ifwacvfinal
\def\assignedStartPage{1} 
\fi

\ifwacvfinal
\usepackage[breaklinks=true,bookmarks=false]{hyperref}
\else
\usepackage[pagebackref=true,breaklinks=true,colorlinks,bookmarks=false]{hyperref}
\fi

\ifwacvfinal
\else
\fi

\begin{document}
\title{Towards Class-Oriented Poisoning Attacks Against Neural Networks}

\author{Bingyin Zhao \quad \quad Yingjie Lao\\
Department of Electrical and Computer Engineering\\
Clemson University, SC, 29634, USA\\
{\tt\small \{bingyiz, ylao\}@clemson.edu}
}
\date{}

\maketitle
\begin{abstract}
Poisoning attacks on machine learning systems compromise the model performance by deliberately injecting malicious samples in the training dataset to influence the training process. Prior works focus on either availability attacks (i.e., lowering the overall model accuracy) or integrity attacks (i.e., enabling specific instance based backdoor). In this paper, we advance the adversarial objectives of the availability attacks to a per-class basis, which we refer to as class-oriented poisoning attacks. We demonstrate that the proposed attack is capable of forcing the corrupted model to predict in two specific ways: (i) classify unseen new images to a targeted ``supplanter'' class, and (ii) misclassify images from a ``victim'' class while maintaining the classification accuracy on other non-victim classes. To maximize the adversarial effect as well as reduce the computational complexity of poisoned data generation, we propose a gradient-based framework that crafts poisoning images with carefully manipulated feature information for each scenario. Using newly defined metrics at the class level, we demonstrate the effectiveness of the proposed class-oriented poisoning attacks on various models (e.g., LeNet-5, Vgg-9, and ResNet-50) over a wide range of datasets (e.g., MNIST, CIFAR-10, and ImageNet-ILSVRC2012) in an end-to-end training setting. 
\end{abstract}

\section{Introduction} \label{sec:intro}
In recent years, machine learning has demonstrated superior performance in various fields including computer vision~\cite{krizhevsky2012imagenet}, natural language processing~\cite{DBLP:journals/corr/abs-1810-04805}, autonomous vehicle~\cite{DBLP:journals/corr/BojarskiTDFFGJM16}, and healthcare~\cite{esteva2017dermatologist}. However, it has also been shown that machine learning models are vulnerable to various types of attacks, including evasion attacks~\cite{DBLP:conf/icml/AthalyeC018,DBLP:journals/corr/GoodfellowSS14,moosavi2016deepfool,DBLP:journals/corr/SzegedyZSBEGF13} backdoor attacks~\cite{DBLP:journals/corr/abs-1712-05526,gu2017badnets,DBLP:conf/globalsip/ClementsL18,DBLP:journals/corr/abs-1806-05768, DBLP:conf/iscas/ClementsL19,DBLP:journals/corr/abs-1912-02771,DBLP:conf/aaai/SahaSP20} and poisoning attacks~\cite{DBLP:conf/icml/BiggioNL12,mei2015using,munoz2017towards,DBLP:conf/malware/ZhaoL18,shafahi2018poison,steinhardt2017certified,DBLP:journals/corr/abs-1906-07773,DBLP:journals/corr/YangWLC17}. Evasion attacks occur at the inference phase, which causes misclassification without altering the model. Backdoor attacks raise misclassifications on specific inputs embedded with certain triggers, which requires access to both training and inference phase to inject and activate backdoor triggers. In contrast, poisoning attacks corrupt the model by only injecting malicious training data in the training phase, without requiring attackers take control of model inputs during inference. The category of attacks has drawn particular attention under the scenario where attackers are able to provide training data (e.g. online repositories).


Prior research on poisoning attacks can be broadly classified into two categories: availability attacks that aim at degrading overall model accuracy (i.e., denial-of-service attacks)~\cite{DBLP:conf/icml/BiggioNL12,DBLP:conf/ecai/XiaoXE12,mei2015using,munoz2017towards,mozaffari2014systematic,jagielski2018manipulating,steinhardt2017certified,DBLP:journals/corr/YangWLC17,DBLP:journals/corr/abs-1906-07773} and integrity attacks that seek to cause misclassification on specific instances (i.e., a targeted image)~\cite{shafahi2018poison,DBLP:conf/icml/ZhuHLTSG19,huang2020metapoison}.
While various capabilities of integrity attacks on deep neural networks (DNNs) have been comprehensively investigated, most prior studies of availability attacks are in a very constrained setting. Poisoning availability attacks had mainly focused on binary classification tasks until~\cite{munoz2017towards} proposed an efficient algorithm for multi-class attack. However, the authors explicitly pointed out that poisoning availability attack against DNN is challenge and the effect of their method is not significant. On the other hand, poisoned data are notoriously hard to craft due to computational complexity of solving the bi-level optimization (see Section~\ref{sec:settings} for details). Moreover, the major adversarial goal of prior works on poisoning availability attack is only limited to degrading the overall accuracy.

Given these limitations, we extend the \textbf{poisoning availability attack} against DNNs to a per-class basis. We advance the adversarial objectives by formulating two attack tasks: (i) forcing the model to classify all new inputs as a targeted class, which is denoted as the \emph{supplanter class} and (ii) corrupting performance of a specific class, which is named as the \emph{victim class}, while retaining the accuracy of other classes. Note that (ii) can be considered as an extension of the targeted poisoning attack that aims to induce the model to make wrong predictions on a victim class. The essential difference is that we minimize the attack impact on non-victim classes simultaneously. We propose a fast and efficient gradient-based framework for poisoned data generation, which reduces the computational complexity and generates more effective poisoned samples. 

\section{Related Work}
Existing literature studied the poisoning availability attack on binary classification tasks against various learning algorithms such as clustering~\cite{biggio2013data}, LASSO~\cite{xiao2015feature}, collaborative filtering~\cite{li2016data}, SVM~\cite{DBLP:conf/icml/BiggioNL12} and logistic regression~\cite{DBLP:journals/corr/abs-1906-07773}. 
The main challenge of the poisoning attack is the generation of effective poisoned data. Prior works developed a series of gradient-based approaches for poisoned data generation, including substituting the inner minimization problem with stationary Karush-Kuhn-Tucker (KKT) conditions~\cite{mei2015using}, approximating the non-convex and non-differentiable models to influence functions~\cite{koh2017understanding} and gradient ascent optimizations~\cite{DBLP:conf/icml/BiggioNL12,li2016data}.
In this work, we approximate the formulated optimization problems to reduce the computational complexity, which will be discussed in Section~\ref{sec:method}. 

The work in~\cite{munoz2017towards} firstly proposed back-gradient optimization and extended poisoning availability attack to multi-class classification. However, the attack is less effective against DNNs.~\cite{DBLP:journals/corr/YangWLC17} expedited the poisoned data generation using generative models and evaluated the effect on MLP and LeNet.~\cite{DBLP:journals/corr/abs-2106-10807} leveraged adversarial examples as poisoned data. Prior works mostly focused on indiscriminately degrading the overall accuracy and disregarded particular prediction error of each class. In this work, we propose algorithms that focus on optimizing the feature information of the most important classes, which is able to achieve the class-oriented adversarial goals as well as facilitate the multi-class poisoning availability attack against DNNs.
\cite{DBLP:conf/nips/FengCZ19} studied an interesting learning problem that is similar to the class-oriented availability attack. However, the essential difference with~\cite{DBLP:conf/nips/FengCZ19} is that it mainly focuses on adding smallest bounded noises to the entire training data whereas our work focuses on generating a portion of the most effective poisoned data.

Poisoning integrity attacks and backdoor attacks have been extensively studied in the literature~\cite{DBLP:journals/corr/abs-1712-05526,gu2017badnets,DBLP:journals/corr/abs-1912-02771,DBLP:conf/aaai/SahaSP20,shafahi2018poison,DBLP:conf/icml/ZhuHLTSG19,huang2020metapoison}. For instance, \cite{DBLP:journals/corr/abs-1712-05526} imposed backdoor data poisoning and caused face recognition systems to misclassify images that contain a ``glasses'' pattern. ~\cite{DBLP:conf/aaai/SahaSP20} proposed hidden trigger backdoor attacks where triggers are more stealthy and imperceptible to human inspection. ~\cite{shafahi2018poison} proposed an integrity attack approach that employed feature collision to cause misclassification on a specific target image and evaluated on ImageNet dataset. ~\cite{DBLP:conf/icml/ZhuHLTSG19} enhanced the transferability of integrity attacks by crafting poisoned images surrounded the targeted image and~\cite{huang2020metapoison} accelerated poisoned data generation of such attacks. However, as we described in Section~\ref{sec:intro}, poisoning availability attack is fundamentally different from poisoning integrity attacks or backdoor attacks. 
Note that poisoning integrity attacks and backdoor attacks are not class-oriented since the adversarial goal is only on selected instances instead of object classes.

\section{Problem Settings}
\label{sec:settings}
\subsection{Poisoning Availability Attack Setting}

We consider the scenario where a DNN is initialized with pre-trained weights and then updated on a poisoned dataset in a full end-to-end fashion. This scenario is one of the most pervasive poisoning attack settings and widely adopted in the state-of-the-art research literature~\cite{munoz2017towards,DBLP:journals/corr/YangWLC17,shafahi2018poison,DBLP:conf/icml/ZhuHLTSG19,DBLP:conf/icml/SchwarzschildGG21} since pre-trained networks are frequently used in real-world applications. To define the problem, let $\boldsymbol{x} \in \mathcal{X}(\mathcal{X} \in \mathbb{R}^d)$ be a $d$-dimensional input and $y \in \mathcal{Y}$ be the corresponding label. The objective of the classification task is to build up the mapping $\mathcal{F}$: $\mathcal{X} \xrightarrow{} \mathcal{Y}$. We denote the parameters of pre-trained base classifier  as $\theta$. The model parameters are updated to $\theta^*$ with the incoming new stream of data: $\theta \xrightarrow{(\boldsymbol{x},y)} \theta^*$.

Poisoning availability attacks are typically formulated as a bi-level optimization problem:
\begin{align}
\mathop{\arg\max}\limits_{\mathcal{D}_{p}}&\sum_{(\boldsymbol{x},y) \in \mathcal{D}_{val}} L\left[\mathcal{F}_{\theta^*} \left( \boldsymbol{x} \right), y, \theta^*\right] \\
\textbf{s.t.}\quad \theta^* \in  \mathop{\arg\min}_{\theta^* \in \Theta}& \sum_{(\boldsymbol{x},y) \in \mathcal{D}_{tr} \cup \mathcal{D}_p} L\left[ \mathcal{F}_{\theta^*} \left( \boldsymbol{x} \right), y,  \theta\right],
\end{align}
where $D_{tr}$ is the clean training dataset, $D_{val}$ is the validation dataset, $D_p$ is the poisoned dataset, $\Theta$ is the possible parameter space, and $L[\cdot]$ is the loss function. The attack aims to find an optimized poisoned dataset, which will be injected into the clean training data for training the benign model and updating parameters. This training process is expressed by the inner minimization. The outer maximization stands for the adversarial objective, which has to be evaluated on the updated parameters found by solving the inner minimization problem.

\subsection{Class-Oriented Adversarial Objectives}
\begin{figure*}[htbp]
    \centering
    \resizebox{1\textwidth}{!}{
    \includegraphics{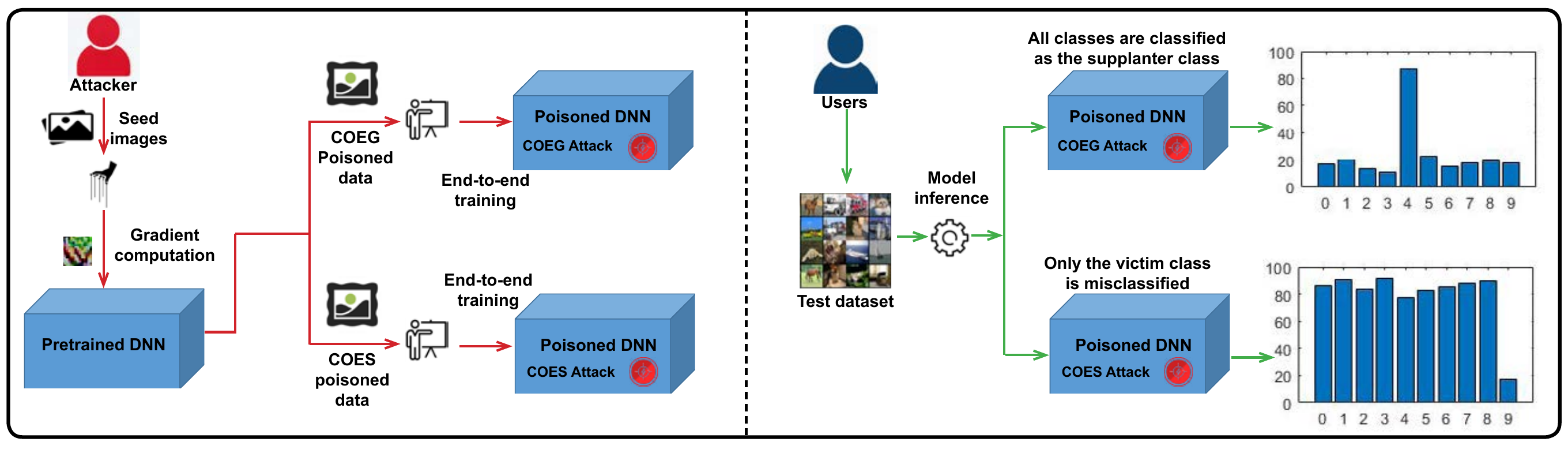}}
    \caption{Class-oriented poisoning availability attacks.}
    \label{task-illustraion}
\end{figure*}

Other than only focusing on maximizing the overall loss, we take the first step towards extending the adversarial objective to a per-class basis, as illustrated in Figure~\ref{task-illustraion}. We impose two new adversarial objectives in addition to the goal of degrading overall accuracy indiscriminately,
which is formulated as two optimization problems accordingly.

\textbf{Problem 1: class-oriented error-generic (COEG)} attack. The goal of 
this attack is to misclassify all or most inputs as a targeted object class, which we named as \emph{supplanter class}. Hence, the overall accuracy will also be degraded. For instance, in Figure~\ref{task-illustraion}, class``4'' is selected as the supplanter class. For a broader real-world example, the adversary can apply the class-oriented error-generic attack to compromise 
a military image classifier to classify all the taken images such as birds and planes as a missile (the supplanter class), raising unnecessary panic or even wrongly activating an anti-missile system. 
The \textbf{COEG} attack problem can be formulated as:

\begin{align}
\mathop{\arg\max}\limits_{ \mathcal{D}_p}&\sum_{(\boldsymbol{x},y) \in \mathcal{D}_{val}} L\left[\mathcal{F}_{\theta^*} \left( \boldsymbol{x} \right), y, \theta^*\right] \label{equ:COEG1}\\
\textbf{s.t.}\quad \theta^* \in \mathop{\arg\min}_{\theta^* \in \Theta}& \sum_{\substack{(\boldsymbol{x},y) \in \mathcal{D}_{tr} \cup \mathcal{D}_p}} L\left[ \mathcal{F}_{\theta^*} \left( \boldsymbol{x} \right), y_s,  \theta\right],\label{equ:COEG2}
\end{align}

where $y_s$ represents the label of the supplanter class.

\textbf{Problem 2: class-oriented error-specific (COES)} attack. The goal is to compromise the classification accuracy only for the inputs from a specific class, which is denoted as \emph{victim class}, while retaining the accuracy of other classes. The bottom right chart in Figure~\ref{task-illustraion} shows the poisoned model behavior where class ``9'' is selected as the victim class. For a broader example again, the adversary can apply the class-oriented error-specific attack to 
the military image classifier such that only missiles (the victim class) will not be correctly classified, resulting in severe security risks. 
The \textbf{COES} attack problem is formulated as:
\begin{align}
\mathop{\arg\max}\limits_{\mathcal{D}_p}& \sum_{\substack{(\boldsymbol{x},y) \in \mathcal{D}_{val} }} L\left[ \mathcal{F}_{\theta^*} \left( \boldsymbol{x} \right), y_v, \theta^*\right] \\
\textbf{s.t.}\quad \theta^* \in \mathop{\arg\min}_{\theta^* \in \Theta}& \sum_{\substack{(\boldsymbol{x},y) \in \mathcal{D}_{tr\cup \mathcal{D}_p}}} L\left[\mathcal{F}_{\theta^*} \left( \boldsymbol{x} \right), y_{\Bar{v}}, \theta\right],  
\label{equ:COES}
\end{align}
where $y_v$ and $y_{\Bar{v}}$ represent the labels of the victim class and non-victim classes, respectively.

\subsection{Class-Oriented Evaluation Metric}

We propose two class-oriented evaluation metrics to assess the performance of class-oriented poisoning attacks. 

\textbf{Change-to-Target (CTT)} rate is designed as an evaluation metric for the \textbf{COEG} attack, which indicates the percentage of images that are classified as a targeted supplanter class ($\mathcal{C}_s$) due to the poisoning attack. CTT rate for a class $\mathcal{C}_k$ is formally defined over a validation dataset $\mathcal{D}_{val}$ as:
\begin{equation}
 CTT(\mathcal{C}_k) = \frac{1}{N_k}\underset{\substack{(\boldsymbol{x_i},y_i) \in \mathcal{D}_{val} \\{y_i=y_k}}} {\sum} \big{(}{\mathcal{F}_{\theta^*}(\boldsymbol{x_{i}})}_{y_s} -{\mathcal{F}_{\theta}(\boldsymbol{x_{i}})}_{y_s}\big{)},
    \label{equ_CTT_perclass}
\end{equation}

\begin{equation}
    \ \mathcal{F}_{\theta}(\boldsymbol{x_{i}})_{y_s} = 
    \begin{cases}
    1 & \text{if  $\mathcal{F}_{\theta}(\boldsymbol{x_{i}}) = y_s$}\ \\
    0 & \text{otherwise}, 
    \end{cases}
    \label{ctt_prediction}    
\end{equation}
where $N_k$ is the total number of images in the class $\mathcal{C}_k$ and $y_k$ is the corresponding categorical label, while $\mathcal{F}_{\theta}$ and $\mathcal{F}_{\theta^*}$ are the model inference results before and after the poisoning attack, respectively. Then, the overall CTT rate over $\mathcal{D}_t$ can be calculated by weighted averaging the CTT rates of all non-supplanter classes:
\begin{equation}
    CTT  =\frac{\sum_{\substack{k=1 \\ k\neq s}}^{K}  \big{(}N_k \cdot CTT(\mathcal{C}_k)\big{)}}{\sum_{\substack{k=1 \\ k\neq s}}^{K} N_k},
    \label{equ_CTT}
\end{equation}
where $K$ is the total number of classes.

\textbf{Change-from-Target (CFT)} rate is specifically used for evaluating the \textbf{COES} attack, which shows the percentage of images from a targeted class are misclassified due to the poisoning attack. Similarly, CFT rate for a class $\mathcal{C}_k$ is defined by Equation~(\ref{equ_CFT}):
\begin{equation}
    CFT(\mathcal{C}_k) =\frac{1}{N_k}\underset{\substack{(\boldsymbol{x_i},y_i) \in \mathcal{D}_{val} \\{y_i=y_k}}} {\sum} \big{(}{\mathcal{F}_{\theta^*}(\boldsymbol{x_{i}})}_{y_k} -{\mathcal{F}_{\theta}(\boldsymbol{x_{i}})}_{y_k}\big{)}.
    \label{equ_CFT}
\end{equation}

\subsection{Threat Model}
We consider a threat model that is consistent with prior works on poisoning availability attacks~\cite{mei2015using,koh2017understanding,munoz2017towards,DBLP:journals/corr/YangWLC17,DBLP:journals/corr/abs-1906-07773}, which assumes the adversary to have the knowledge of the learning algorithm, hyper-parameters and clean training data. The attack is performed in an end-to-end training setting on a benign model. The adversary is able to inject crafted poisoned data and assign labels to the training dataset, which also holds the same assumption as the prior literature~\cite{DBLP:conf/icml/BiggioNL12,li2016data,burkard2017analysis,munoz2017towards,DBLP:journals/corr/abs-1906-07773}. However, it is worth mentioning that our proposed approach also works under a more practical and strict scenario where the adversary only has limited knowledge of the model where only model architecture and pre-trained weights are required, while neither the learning algorithm nor the original training dataset are assumed to be known to the adversary.


\section{Class-Oriented Poisoning Attack Methods}\label{sec:method}
\subsection{COEG Attack}
Intuitively, we expect retraining images with the label of the supplanter class, similar to the flipped-label poisoning attack~\cite{DBLP:conf/ecai/XiaoXE12}, would have the potential to shift predictions of all classes towards the supplanter class. This is the most straightforward approach that does not even require crafting poisoned data. However, our experimental results demonstrate that such attacks, when applied to multi-class classification tasks, neither effectively degrade the overall accuracy nor achieve the class-oriented adversarial goal of the COEG attack for the neural network models (see Section~\ref{sec:exp}). 

Alternatively, we can directly solve Equations~(\ref{equ:COEG1})-(\ref{equ:COEG2}) to get poisoned data through gradient ascent. However, for non-convex learning problem such as in DNN, it is difficult to compute $\frac{\partial \theta ^*}{\partial D_p}$. To simplify the poisoned data generation and improve the effectiveness of poisoning towards the supplanter class, we develop a novel and efficient method to craft poisoned data. We leverage the fact that the probabilities assigned to other object classes of a well-trained model reveal how much feature information of these incorrect classes is associated with the corresponding image by the model~\cite{DBLP:journals/corr/HintonVD15}. Thus, we hypothesize that if an input image only contains feature information of its ground-truth class, training such an image with the supplanter class label will force the model to expand the decision boundary of the supplanter class to the maximum degree. 

We follow the direction of prior works, which exploit the logit outputs to distill knowledge of neural networks~\cite{DBLP:journals/corr/HintonVD15}, catch features~\cite{ilyas2019adversarial}, and defend against adversarial examples~\cite{DBLP:journals/corr/abs-1803-06373}, to control the feature information through the logit outputs. Our algorithm starts with a seed image $\boldsymbol{x_o}$ that is arbitrarily picked from any class other than the supplanter class, and then attempts to retain the feature information associated with the ground-truth class and reduce the feature information of other classes by enlarging/dwindling the corresponding logit outputs. We here denote $f(\cdot)$ as the logit output function of the neural network and $f_{y_k}$ as the corresponding logit to the categorical label $y_k$. The objective of our algorithm can be expressed by the following 
minimization objective function:
\begin{equation}
        L = \lambda_k\cdot L_{\sum{f_{y_k}}} - L_{f_{y_o}} ,
\end{equation}
\begin{equation}
\begin{aligned}
        &L_{f_{y_o}}  = f_{y_o}(\boldsymbol{x}),\\
        &L_{\sum{f_{y_k}}}  = \sum_{\substack{k =1\\k \neq o}}^{K}{f_{y_k}(\boldsymbol{x})}.
\end{aligned}
        \label{objective_function}
\end{equation}
where $\boldsymbol{x}$ is the image being optimized, which is initialized with the seed image $\boldsymbol{x_o}$. $y_o$ is the corresponding categorical ground-truth label, $f_{y_o}(\boldsymbol{x})$ is the logit output of the ground-truth class, $f_{y_k}(\boldsymbol{x})$ is the logit output of each other classes, and finally $\boldsymbol{x_p}$ stands for the optimized poisoned image. $\lambda_k$ is used to control the importance of loss terms. We attempt to maximize $L_{f_{y_o}}$ and minimize $L_{\sum{f_{y_k}}}$ simultaneously. Alternatively, from the perspective of entropy, we expect such optimization would also reduce the following information entropy $H[\cdot]$:
\begin{equation}
    H\big{[}\sigma\big{(}f(\boldsymbol{x_p})\big{)}\big{]}=-\sum_{k=1}^{K}p_k\cdot log(p_k) \xrightarrow{} 0,
    \label{entropy}
\end{equation}
where $p_k$ is the probability for each class that is converted from the logit $f_{y_k}$ using \emph{softmax}.

\begin{algorithm}[tb]
  \caption{COEG Poisoned Data Generation}
  \label{algo:poisoned data generation}
\begin{algorithmic}
  \STATE {\bfseries Input:} $\boldsymbol{x_o}$: seed image, $y_o$: seed image label, $y_s$: supplanter class label, $T$: max number of optimization iterations, hyper-parameters $\lambda$, $\epsilon$
  \STATE {\bfseries Output:} poisoned image $\boldsymbol{x_p}$, poisoned label $y_p$
  \STATE Initialize: $\boldsymbol{x_{p_0}} =  \boldsymbol{x_o} - \epsilon \cdot sign\Big{(}\nabla_{\boldsymbol{x_o}}\big{(} \lambda\cdot L_{f_{y_s}} - L_{f_{y_o}}\big{)}\Big{)} $
  \WHILE{$t < T$}
  \STATE  Compute the gradient: $\nabla = \nabla{\boldsymbol{x_{p_t}}}\big{(}\lambda\cdot L_{f_{y_s}} - L_{f_{y_o}}\big{)}$
  \STATE Update the image: 
    {Clip}\{$\boldsymbol{x_{p_{t+1}}} = \boldsymbol{x_{p_t}} - \epsilon \cdot sign(\nabla)$\}
  \IF{$f_{y_s}(\boldsymbol{x_{p_{t+1}}})$ $>$ $f_{y_s}(\boldsymbol{x_{p_{t}}})$ \textbf{or} $f_{y_o}(\boldsymbol{x_{p_{t+1}}})$ $<$ $f_{y_o}(\boldsymbol{x_{p_{t}}})$}
  \STATE break
  \ENDIF
  \ENDWHILE
  \STATE \textbf{Assign} $y_p = y_s$
  \STATE \textbf{Return} $\boldsymbol{x_p}, y_p$ 
\end{algorithmic}
\end{algorithm}

However, solving the optimization problem can be computationally expensive, especially for large-scale datasets that have hundreds or thousands of classes. Based on the facts that (i) the classification is determined by the largest probability and (ii) only the supplanter class is the target, we consider an approximation that simplifies the task to retain feature information of the ground-truth class and eliminate feature information of the supplanter class. In other words, we only focus on the two most important classes instead of all classes:  

\begin{equation}
\begin{aligned}
        &L = \lambda\cdot L_{f_{y_s}} - L_{f_{y_o}},\\
        &L_{f_{y_s}}  = f_{y_s}(\boldsymbol{x}).
        \label{objective_function_simplified}
\end{aligned}
\end{equation}

By solving the minimization using gradient descent, the poisoned image $\boldsymbol{x_p}$ is updated through one backward pass:
\begin{equation}
    \boldsymbol{x_p} = \boldsymbol{x_o} - \epsilon \cdot sign\Big{(}\nabla_{\boldsymbol{x_o}}\big{(}\lambda\cdot L_{f_{y_s}} - L_{f_{y_o}}\big{)}\Big{)},
    \label{poisoned_data_update}    
\end{equation}
where $\epsilon>0$ is the change rate. This update step can also be executed for several rounds to further enhance the poisoning effect. Algorithm~\ref{algo:poisoned data generation} presents the details of poisoned data generation for the COEG attack.

    

    
    
    
    
    
   


\subsection{COES Attack}

The COES attack is fundamentally more challenging than the COEG attack. It requires not only compromising the accuracy of the targeted victim class but also maintaining the performance of non-victim classes, i.e.,  a high CFT rate for the victim class while low CFT rates for all the other classes. However, as observed in our single instance attack experiment as well as prior poisoning availability attacks~\cite{park2017resilient}, poisoning with data from only one class will shift the distribution of other classes to some extent. Hence, a set of poisoned data from more than a single class is necessary to achieve this adversarial goal. Intuitively, simply training images from all classes except the victim class may achieve this adversarial goal. However, such methods are inefficient in the end-to-end retraining scenario where limited training data and small learning rates are usually applied, as they always require much more training data and training epochs for poisoning.

    
  






To this end, we propose another gradient-based algorithm for the COES attack. The entire procedure of our poisoned data generation is presented in Algorithm~\ref{algo:poisoned data generation_t2}. We craft the poisoned dataset as follows: (i) pick a same number of arbitrary images from each class, (ii) enlarge/dwindle feature information of the corresponding classes for each image, as detailed in Algorithm~\ref{algo:poisoned data generation_t2}, and (iii) assign the ground-truth labels to the non-victim classes and the targeted label $y_p$ to the victim class. Specifically, for images from the victim class, we apply the same operations as in Algorithm~\ref{algo:poisoned data generation}, while we only increase the feature information of their ground-truth classes for images from other classes. The objective of COES attack can be expressed as:
\begin{equation}
\ L = 
    \begin{cases}
    \lambda\cdot L_{f_{y_s}} - L_{f_{y_o}} , & \text{if  $\boldsymbol{x_o} \in \mathcal{C}_v$}\ \\
    L_{f_{y_o}} , & \text{otherwise} 
    \end{cases}
    \label{objective_function_3}    
\end{equation}

Similarly, the poisoned images $\boldsymbol{x_p}$ are updated through a backward pass:
\begin{equation}
\hspace{-0.6cm}
    \begin{cases}
    \boldsymbol{x_o} - \epsilon \cdot sign\Big{(}\nabla_{\boldsymbol{x_o}}\big{(}\lambda\cdot L_{f_{y_s}} - L_{f_{y_o}}\big{)}\Big{)}, &\text{if  $\boldsymbol{x_o} \in \mathcal{C}_v$}\ \\
    \boldsymbol{\boldsymbol{x_o}} + \epsilon \cdot sign\Big{(}\nabla_{\boldsymbol{x_o}}\big{(}L_{f_{y_o}}\big{)}\Big{)},  &\text{otherwise} 
    \end{cases}
    \label{poisoned_data_generation_2} 
\end{equation}

\begin{algorithm}[tb]
  \caption{COES Poisoned Data Generation}
  \label{algo:poisoned data generation_t2}
\begin{algorithmic}
  \STATE {\bfseries Input:} $\boldsymbol{x_k} \in \boldsymbol{X}$: the set of seed images,\\
  \STATE  {$y_k \in Y$}: the set of labels associated to $X$, \\$y_p$: poisoned label, $T$: max number of iterations, hyper-parameters $\lambda$, $\epsilon$
  \STATE {\bfseries Output:} poisoned dataset $\boldsymbol{X_p}$, poisoned label set $Y_p$
  \IF{$\boldsymbol{x_k}$ $\in$ $\mathcal{C}_v$}
  \STATE \textbf{Apply} Algorithm 1
  \STATE \textbf{Assign} $y_k = y_p$
  \ELSE
  \STATE  Initialize: $\boldsymbol{x_{p_0}} = \boldsymbol{\boldsymbol{x_o}} + \epsilon \cdot sign\Big{(}\nabla_{\boldsymbol{x_o}}\big{(}L_{f_{y_o}}\big{)}\Big{)}$
  \WHILE{$ t < T$}
  \STATE Compute the gradient: $\nabla = \nabla{\boldsymbol{x_{p_t}}}\big{(}L_{f_{y_o}}\big{)}$
  \STATE Update the image: Clip\{$\boldsymbol{x_{p_{t+1}}} = \boldsymbol{x_{p_t}} + \epsilon \cdot sign(\nabla)$\}
  \ENDWHILE
  \ENDIF
  \STATE \textbf{Assign} $\boldsymbol{X_p} = \boldsymbol{X}$ ; $Y_p = Y$
  \STATE \textbf{Return} $\boldsymbol{X_p}$, $Y_p$
\end{algorithmic}
\end{algorithm}

\section{Experiments} \label{sec:exp}
\subsection{Experimental Settings}
As described above, we consider a similar setting as in prior works~\cite{DBLP:journals/corr/YangWLC17,shafahi2018poison}. We apply our proposed class-oriented poisoning availability attacks to multi-class image classification tasks using three widely-used datasets (MNIST, CIFAR-10, and ImageNet-ILSVRC2012) against popular neural network models (LeNet-5~\cite{lecun1998gradient}, Vgg-9~\cite{DBLP:journals/corr/SimonyanZ14a}, and ResNet-50~\cite{he2016deep}, respectively). Model details are presented in the appendix. For comparison with prior works, we implement the flipped-label (FL) attack~\cite{DBLP:conf/ecai/XiaoXE12} and the direct gradient method (DGM)~\cite{DBLP:journals/corr/YangWLC17} as the baseline attacks for MNIST and CIFAR-10. We also examine our poisoning availability attacks on ImageNet against a ResNet-50 model, which we hope to serve as a baseline comparison for future works. To illustrate the effectiveness of our methods under a fair comparison, we minimize the impact of significant model shifts due to large learning rates by setting the initial learning rate of the poisoning attack close to the final learning rate of the base model and applying the same decay strategy during each attack. All networks are implemented with TensorFlow and experiments are run on NVIDIA Tesla V100 GPUs.

\subsection{Experimental Results of COEG Attack}
\subsubsection{Single instance attack.} 
\begin{figure}[htbp]
    \centering
    \resizebox{0.47\textwidth}{!}{
    \includegraphics{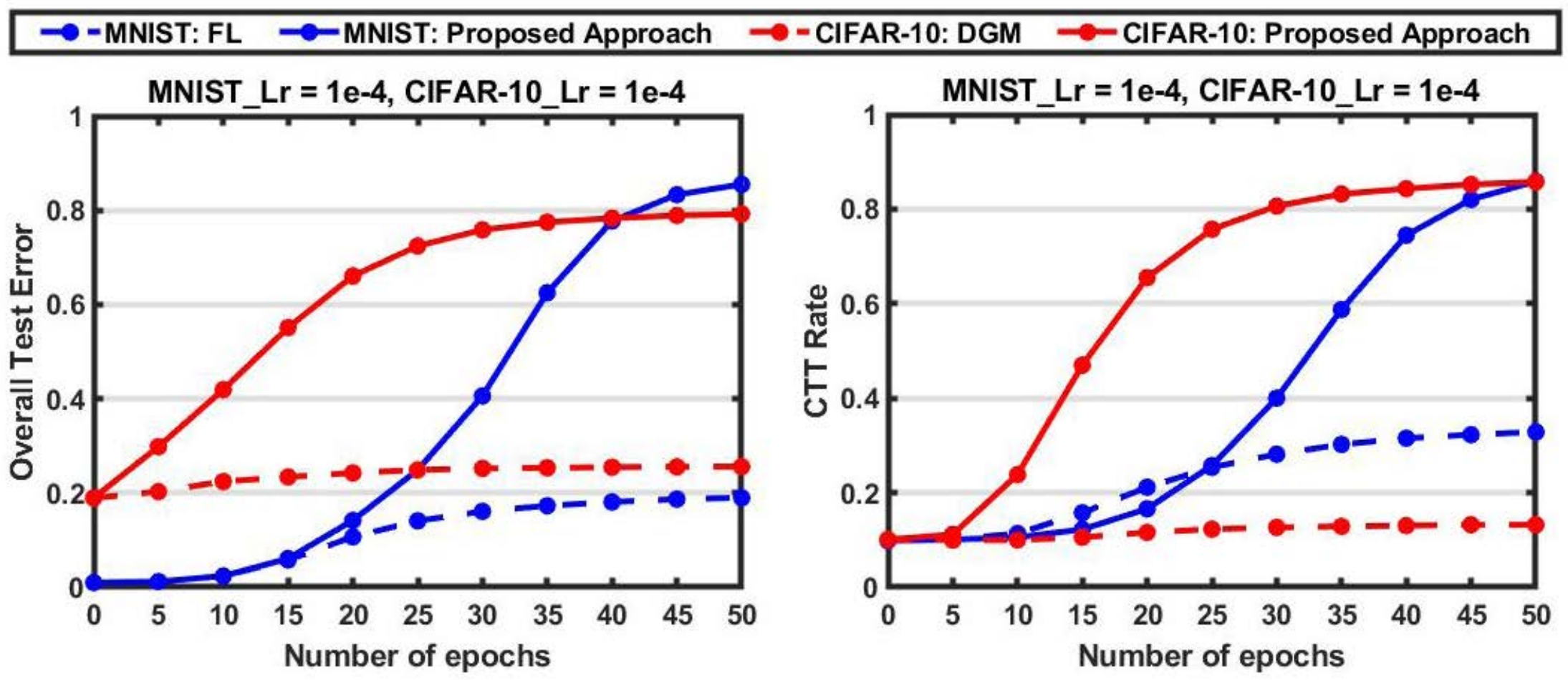}}
    \caption{Error and CTT rate comparison of single instance attack. Classes `4' and `deer' are selected as the supplanter class for MNIST and CIFAR-10, respectively.}
    \label{Error-CTT}
\end{figure}

We first evaluate the effectiveness of our COEG attack under the single instance attack setting~\cite{DBLP:journals/corr/YangWLC17} that implements the poisoning attack by only training with a sole poisoned point. Hyper-parameters in Algorithm 1 are set as: $\lambda =1, \epsilon = 0.3$ for MNIST and CIFAR-10.
We present the results of MNIST and CIFAR-10 along with baseline methods (FL and DGM) in Figure~\ref{Error-CTT}. 

\begin{table}[htbp]
\centering
\caption{Comparison of CTT and accuracy before and after the COEG attack on ImageNet.}
\scalebox{0.76}{
  \begin{tabular}{ c | c  c  c }
      \toprule[1pt]
         & Top-1 Accuracy & Top-5 Accuracy & CTT\\
      \midrule[1pt]
    Vanilla ResNet-50  & 74.87\% & 92.02\% & --\\
    Poisoned by Our Attack  & {6.73\%} & 15.00\% & {85.60\%}\\

      \bottomrule[1pt]
  \end{tabular}
  }

\label{resnet-CTT}

\end{table}

It can be seen that our proposed attack is highly effective in degrading the overall model accuracy and increasing the CTT rate for the supplanter class. For example, we increase the test error from 20\% to $\sim$70\% and the CTT rate from 10\% to $\sim$60\% on the CIFAR-10 dataset within 20 epochs. Besides, with the increase of training epochs, our method can still consistently achieve higher test errors and CTT rates than baseline attacks, which only increase the test error and CTT rate by $\sim$8\% and $\sim$0.5\%, respectively, even after 50 training epochs. 
The result for ResNet-50 on ImageNet with hyper-parameters $\lambda =1, \epsilon = 0.5$ is presented in Table~\ref{resnet-CTT}. Our proposed method achieves a CTT rate of 85.60\% within 20 epochs in the single instance attack. One may argue that the success is due to the bias yielded from training with a single data point. However, for a robust benign model, the accuracy drop and CTT rate of DGM and FL attacks using the same training strategy with a single instance are limited, as shown in Figure~\ref{Error-CTT}, while our attack achieves much better performance. 

\begin{figure}[htbp]
    \centering
    \resizebox{0.48\textwidth}{!}{
    \includegraphics{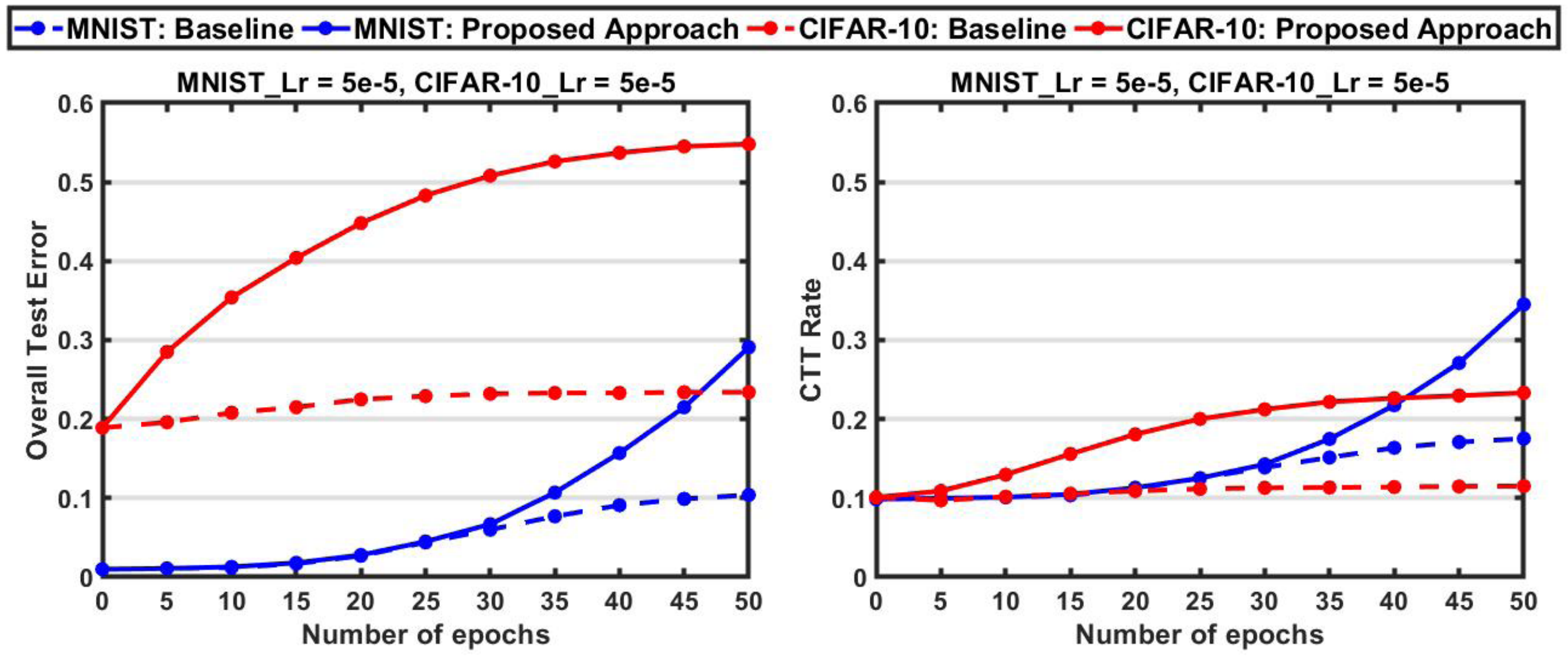}}
    \caption{Error and CTT rate comparison of single instance attack with smaller learning rate at $5 \times 10^{-5}$. Classes `4' and `deer' are selected as the supplanter class for MNIST and CIFAR-10. Seed images are from class `6' and `frog', respectively.}
    \label{Error-CTT-small-lr}
\end{figure}

For scenarios where the adversary has no knowledge of the training process, it is also essential to study the impact of different learning rates on the attack. Our conclusion is consistent with prior work~\cite{DBLP:journals/corr/abs-1808-08994} that lower learning rates yield less effectiveness for the attack. However, our method still outperforms the baseline attacks when the learning rate is low. We consider a practical adversary model that attackers can only inject the poisoned data but have no knowledge of the learning algorithms and hyper-parameters. To this end, it is worth studying the effect of poisoning attacks with different hyper-parameter settings. Since we have achieved a superior effect with a higher learning rate in Figure 2, we further study the impact of a lower learning rate for the single instance attack. We experiment on both MNIST and CIFAR-10 datasets with a learning rate of $5 \times 10^{-5}$ and use the same baseline attacks for comparison. We find that a lower learning rate reduces the effect of the poisoning attack in this setting, the alteration of the decision boundary provided from each poisoned sample is decreased. However, while it requires more attack iterations to arrive at the maximum poisoning effect, our method still outperforms the baseline attacks when the learning rate is lower, as indicated in Figure~\ref{Error-CTT-small-lr}.

\textbf{Attack with a set of poisoned data.} A more general poisoning attack scenario would allow the attacker to inject a fraction of poisoned data into the clean training dataset, which indeed is a setting often adopted in prior studies~\cite{shafahi2018poison,munoz2017towards,DBLP:journals/corr/abs-1906-07773}. 
We evaluate the effect of our attack on CIFAR-10 by using 1000 images for training and 9000 images for testing. The number of poisoned samples are controlled by the fraction parameter $\alpha$. For example, when $\alpha = 0.1$, 100 images are poisoned and 900 images remain clean. In the experiment, class ``airplane'' is set as the supplanter class. The poisoning attack is conducted at learning rate of $1 \times 10^{-5}$ over 20 epochs and batch size of 128. We also implement the FL attack as our baseline. 
 \begin{figure}[htbp]
    \centering
    \resizebox{0.47\textwidth}{!}{
    \includegraphics{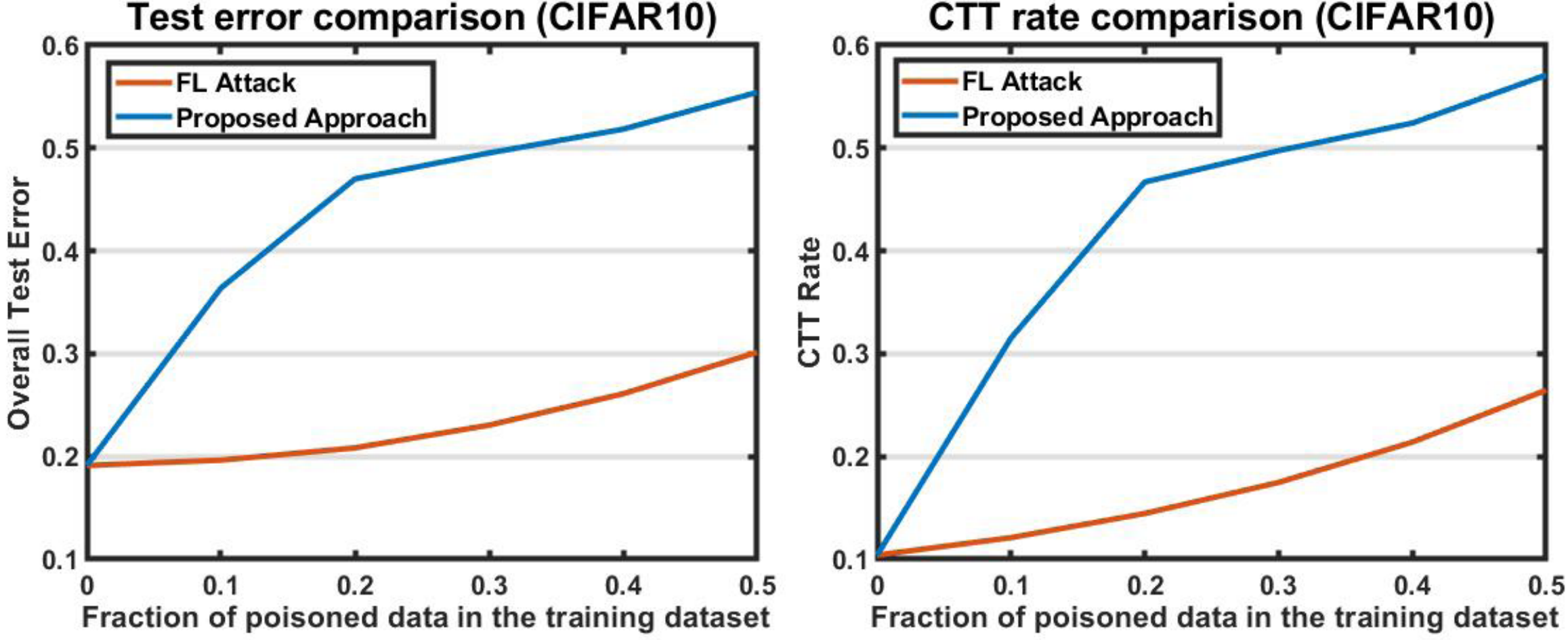}}

    \caption{Error and CTT rate comparison of general poisoning attack with different poisoning fraction.}
    \label{percentage-ctt-error}
\end{figure}

\begin{figure}[htbp]
    \centering
    \resizebox{0.48\textwidth}{!}{
    \includegraphics{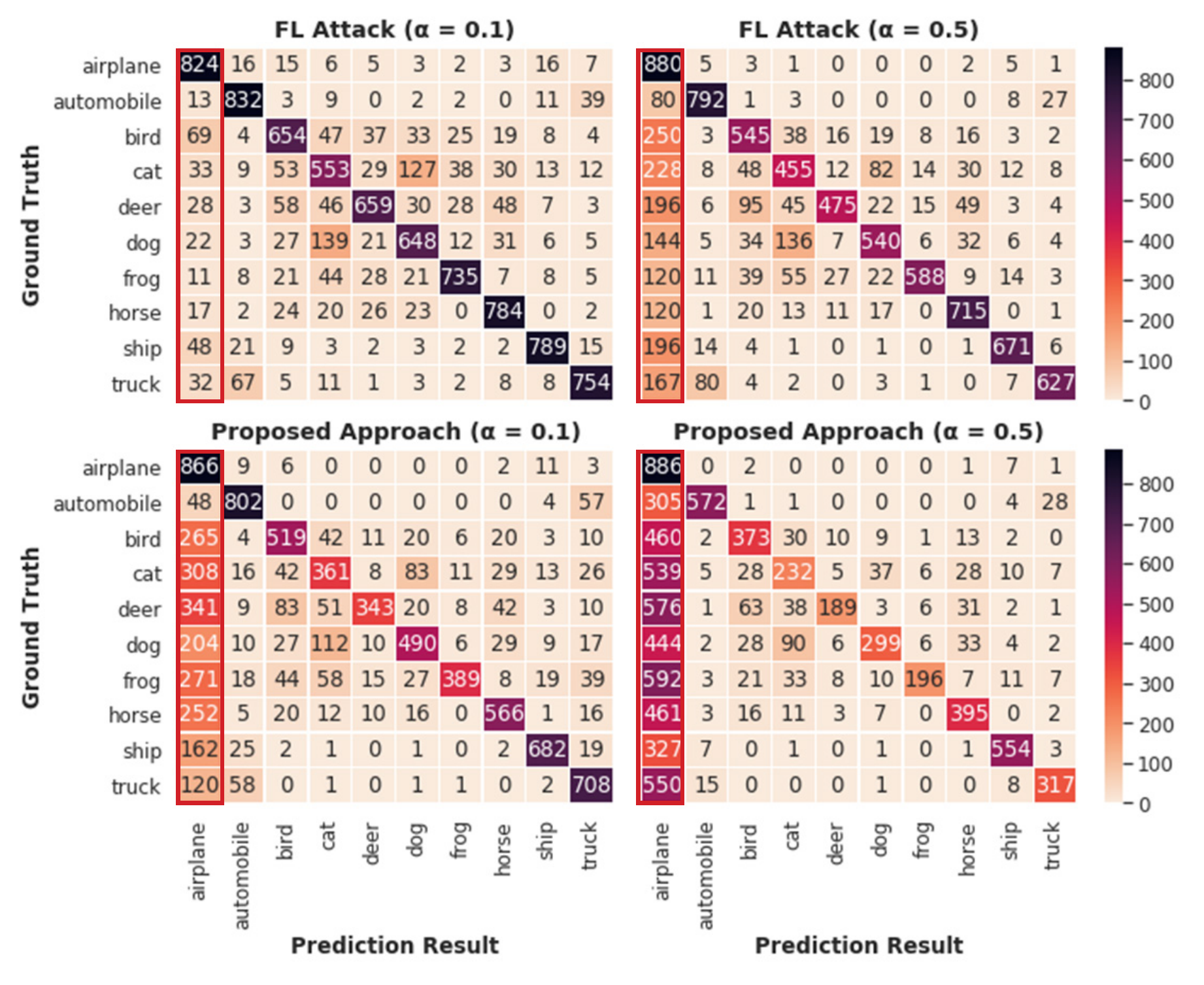}}
    \caption{Confusion matrix with different poisoning fractions. Class ``airplane'' is selected as the supplanter class. Darker color indicates larger change into the supplanter class.}
    \label{heatmap}
\end{figure}

As shown in Figure~\ref{percentage-ctt-error}, our attack outperforms the FL attack for all the $\alpha$ values. We achieve over 50\% for both CTT rate and test error, which are 25\% higher than those of the baseline attack. Note that the proposed attack not only achieves an overall better CTT rate but also performs better on a per-class basis, as indicated in Figure~\ref{heatmap}. The first column of each confusion matrix represents the number of test images classified as the supplanter class after the poisoning attack. Our approach has a darker color (higher CTT rate) for every class than the baseline attack. Another interesting observation is that some classes, e.g., ``automobile'' and ``ship'', are hard to change towards the supplanter class. This is possibly due to the structural similarity of these classes are less distinguishable to the targeted supplanter class. A recent study finds that different seed/target class pairs and training set size may have significantly different poisoning effect~\cite{DBLP:conf/icml/SchwarzschildGG21}. We also examine our attack's performance from this aspect by experimenting with different seed/target pairs, whose results are presented in the appendix. We find that our attack is particularly effective for smaller training set size and outperforms the FL attack with all the seed/target pairs.

\begin{figure}[htbp]
    \centering
    \resizebox{0.48\textwidth}{!}{
    \includegraphics{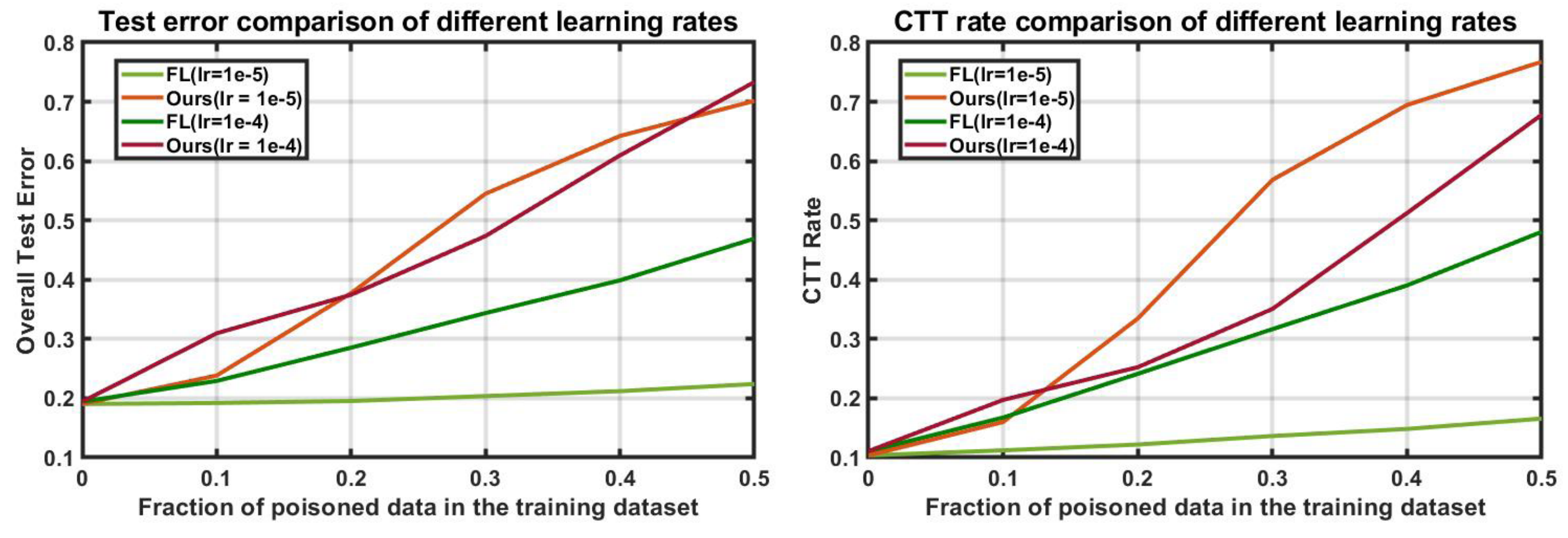}}
    \caption{Comparison of different learning rates for the attack with a set of poisoned data.}
    \label{percentage_error_ctt_larger_lr_500_training_data}
\end{figure}

We also study the impact of different learning rates for the general poisoning attack scenario with more poisoned data. Since we achieve the best performance with a dataset size of 500, we keep the same size for this experiment. We increase the learning rate from $1 \times 10^{-5}$ to $1 \times 10^{-4}$ and keep the remaining settings the same as the previous experiments. Experiment results are shown in Figure~\ref{percentage_error_ctt_larger_lr_500_training_data}. For the FL attack, both the test error and CTT rate are proportional to the learning rate. In contrast, the proposed attack with a lower learning rate achieves a higher test error for $\alpha = 0.2 \sim 0.4$, while yielding a higher CTT rate for $\alpha > 0.1$. This may be attributed to the fact that higher learning rate also amplifies the impact of clean data and partially offsets the effect of poisoned data. Despite this, our proposed attack still achieve better performance compared to the baseline attack in both criteria.

\textbf{Comparison with existing works.}
We compare our approach with state-of-the-art poisoning availability attacks. Since prior works are not class-oriented, we only compare the accuracy drop. Taking the different experimental settings such as victim model and model accuracy into account, we refer to the best results reported in these papers for fair comparison. Also note again that this paper, to the best of our knowledge, is the first work to evaluate poisoning availability attack on ImageNet. So we only compare the performance on MNIST and CIFAR-10. As shown in Table~\ref{comparison_accuracy}, our proposed approach shows superior adversarial capability.

\begin{table}[htbp]
\centering
\caption{Accuracy drop comparison with prior works.}
\scalebox{0.9}{
\begin{tabular}{c|c|c|c}
\hline
\textbf{Approach}      & \textbf{Dataset} & \textbf{\begin{tabular}[c]{@{}c@{}}Accuracy\\ Drop\end{tabular}} & \textbf{\begin{tabular}[c]{@{}c@{}}Victim \\ Model\end{tabular}} \\ \hline
\cite{munoz2017towards} & MNIST            & $\sim$10\%                                                       & LR classifier                                                    \\
\cite{DBLP:journals/corr/YangWLC17}           & MNIST            & $\sim$80\%                                                       & 2 layers NN                                                      \\
\textbf{Ours}          & MNIST            & $\sim$80\%                                                       & LeNet                                                            \\
\cite{DBLP:journals/corr/abs-1906-07773} & CIFAR-10         & $\sim$12\%                                                       & DNN                                                              \\
\textbf{Ours}          & CIFAR-10         & $\sim$30\%                                                       & DNN                                                              \\ \hline
\end{tabular}
}
\label{comparison_accuracy}
\end{table}

\subsection{Experimental Results of COES Attack}
Since the adversarial goal of the COES attack is to subvert only one class without degrading the performance of other classes, there are two important metrics for this task: 1) CFT rate of the victim class should be as high, and 2) CFT rates of the non-victim classes should be as low as possible. Note that the highest achievable CFT rate is upper-bounded by the accuracy of the victim class in the base model. Since the single image attack naturally contradicts to the CEOS adversarial goal (as explained in \emph{Methods Section}), we only consider the general poisoning attack and keep the same training settings and hyper-parameters as in the COEG attack. We present the CFT rate of each class for poisoning attacks on CIFAR-10 in Table~\ref{CFT_CIFAR10_classes}, where class ``truck'' is selected as the victim class and ``airplane'' is selected as the poisoned label. $\alpha=0.5$ is used. We also evaluate the effect of different values of $\alpha$ in the appendix. Two different types of FL attacks are implemented for comparison: FL-1 flips the label of all poisoned images; FL-2 only flips the label of poisoned images from the victim class.
\begin{table}[hbtp]
\centering

\caption{CFT rate of each CIFAR-10 class by poisoning.}

\scalebox{0.73}{
    \begin{tabular}{c|c c c c c}
     \toprule 
     Class / Attack& airplane & automobile & bird & cat & deer \\ 
    \midrule
     FL-1 & -3.90\% & 3.87\% &16.80\% & 16.20\% & 32.65\% \\
     FL-2 & -3.01\% & -1.44\% & -2.05\% & -2.29\% & 2.57\% \\
     \textbf{Ours}  & -7.07\% & 0.88\% & -0.41\% & 2.29\% & 5.27\% \\
     \midrule
      Class / Attack& dog & frog  & horse & ship & \cellcolor[HTML]{C0C0C0}truck(victim)\\ 
     \midrule
     FL-1 & 19.09\% & 22.44\% & 8.34\% & 15.45\% &\cellcolor[HTML]{C0C0C0}{18.42}\% \\
     FL-2  & 2.67\% & -0.59\% & -0.11\% & -0.78\% & \cellcolor[HTML]{C0C0C0}{8.01}\% \\
     \textbf{Ours}   & 3.32\% & -0.47\% & 0.11\% & 0.22\% & \cellcolor[HTML]{C0C0C0}{51.14}\% \\
     \bottomrule

    \end{tabular}
    }
     \label{CFT_CIFAR10_classes}
\end{table}

It can be seen that FL-2 is able to keep the CFT rates of all non-victim classes relatively low; however, the CFT rate of the victim class is only 8.01\%. While the FL-1 attack achieves a CFT rate of 18.42\% for the victim class, it also largely increases the CFT rates of non-victim classes, which indeed verifies the difficulty of achieving the class-oriented adversarial goal by using prior poisoning attack methods. Compared to the baseline attacks, our proposed approach can effectively increase the CFT rate of the victim class to 51.14\%, which significantly surpasses the performance of both baseline attacks. Meanwhile, our method is able to retain the CFT rates of the non-victim classes to be less than 5.27\%. In some cases, the proposed approach even improves the accuracy of some non-victim classes, which is indicated by the negative CFT rates in our experimental results. Such performance is beyond our expectation, which however does not contradict the adversarial objective of COES attack, i.e., only degrading the accuracy of the victim class. Similar to the COEG attack, we find that the proposed COES attack method is more resilient to the variation of learning rate than baseline attacks. We present these results in the appendix. 

\vspace{0.2cm}

\begin{figure}[htbp]
     \centering
     \resizebox{0.48\textwidth}{!}{
     \includegraphics{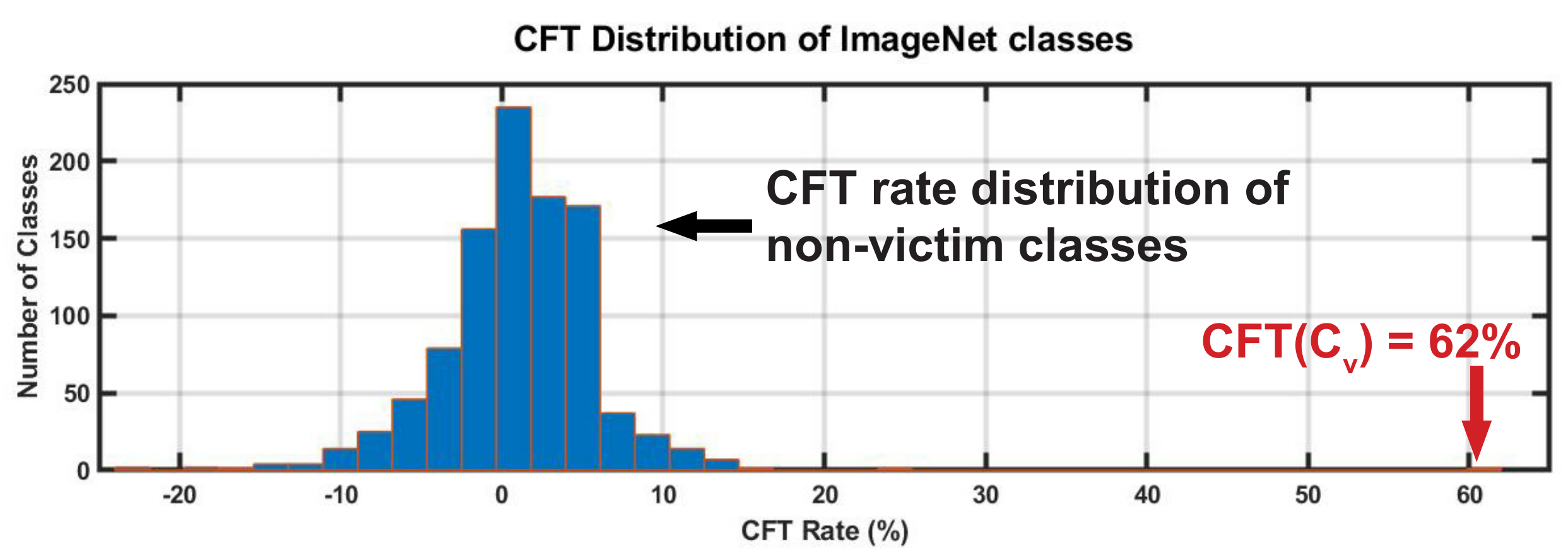}}
     \caption{The CFT rate distribution of ImageNet classes by the COES attack.}
     \label{fig:CFT_resnet}
 \end{figure}

\vspace{0.1cm}
For ImageNet, we inject 100 poisoned images into 1000 clean training images ($\alpha=0.1$). Due to the large number of object classes, we present the distribution of the final CFT rates, as shown in Figure~\ref{fig:CFT_resnet}. Our method  achieves a CFT rate of 62\% for the victim class, while successfully maintaining low CFT rates for non-victim classes. Compared to the performance on CIFAR-10, although we still accomplish the adversarial goal, we find our attack is slightly less effective on ImageNet (i.e. CFT of non-victim classes are more difficult to control). A possible rationale behind this phenomena is that it becomes harder to completely decouple the feature information of one class from all other classes during the poisoning, with the number of classes scaling up. On the other hand, our experimental results also reveal the importance of studying poisoning availability attack on large-scale/dimensional dataset, which lacks a systematic study yet in the existing literature.

\section{Possible Defenses}
Since the main objective of this paper is to extend the adversarial capability of poisoning availability attack to a per-class basis on deep neural networks, we expect the proposed attacks to have similar performance as prior poisoning availability attacks in general when evaluated under possible defenses.

Data sanitization is a defensive technique against poisoning attack that works by distinguishing and removing outliers (poisoned data) from the training dataset~\cite{cretu2008casting,rubinstein2009antidote}. However, it has been shown that a broad range of data sanitization can be easily compromised or bypassed~\cite{DBLP:journals/corr/abs-1811-00741}. Therefore, we can also leverage such techniques for our proposed attacks to evade detection. In fact, in most recent works on poisoning attacks, data sanitization is no longer considered a certified defensive strategy~\cite{DBLP:journals/corr/abs-1906-07773,jagielski2018manipulating,DBLP:journals/corr/abs-1808-08994}.

Alternatively, a possible countermeasure is to periodically check the accuracy and/or loss of the learning models~\cite{munoz2017towards,DBLP:journals/corr/YangWLC17}. Although expensive in terms of cost and time, these approaches are intuitively effective based on the fact that poisoning availability attack aims at degrading the accuracy. Since the poisoned data are tailored to influence the learning model's training process maliciously, we suggest exploiting averaged stochastic gradient classifier~\cite{DBLP:journals/corr/abs-1808-08994} and combinational models such as bagging~\cite{li2016data}, where the classification results are no longer dependent on a single model, to defend against the poisoning attack. However, the overhead for deploying multiple classifiers should also be carefully considered. 






\section{Conclusions}
This paper introduced the concept of class-oriented poisoning attack. We formulated two attack problems, i.e., ``COEG'' and ``COES'', which seek to compromise the model behavior on a per-class basis. Accordingly, we defined two new metrics to evaluate the performance of poisoning attacks at the class level. Our proposed gradient-based algorithms successfully achieved the class-oriented adversarial objectives through manipulating the feature information in images for poisoned data generation. The effectiveness of the proposed methods is comprehensively studied in our experiments. 
\section*{Acknowledgment}
This work is partially supported by the National Science Foundation award 2047384.

\newpage
{\small
\bibliographystyle{ieee_fullname}
\bibliography{egbib}
}

\clearpage
\setcounter{section}{0}





%
 
\def\wacvPaperID{1268} 

\wacvfinalcopy 

\ifwacvfinal
\def\assignedStartPage{11} 
\fi








\section*{Appendix}

\setcounter{figure}{5} 
\setcounter{table}{2}
\section{Neural Network Architectures}
\subsection{LeNet}
The pre-trained LeNet architecture for the single instance attack on the MNIST dataset is summarized in Table~\ref{Lenet-architecture}. Before the poisoning attack, the network is trained with the Adam optimizer at the learning rate of $1 \times 10^{-4}$ for 80 epochs on clean data and achieves a test accuracy of 99\%.

\begin{table}[hbtp]
\caption{Network Architecture for the MNIST experiments}
\scalebox{0.72}{
\begin{tabular}{|c|c|c|c|c|c|c|}
\hline
\multicolumn{2}{|c|}{Layer}                                    & \begin{tabular}[c]{@{}c@{}}Number\\of Filter\end{tabular} & Size  & \begin{tabular}[c]{@{}c@{}}Kernel\\ Size\end{tabular} & Stride & \begin{tabular}[c]{@{}c@{}}Activation\\ Function\end{tabular} \\ \hline
Input  & Image                                                 & -                                                          & 32x32x1 & -                                                     & -      & -                                                             \\ \hline
1      & Convolution                                           & 6                                                          & 28x28 & 5x5                                                   & 1      & ReLU                                                          \\ \hline
2      & \begin{tabular}[c]{@{}c@{}}Max Pooling\end{tabular} & 6                                                          & 14x14 & 2x2                                                   & 2      & -                                                             \\ \hline
3      & Convolution                                           & 16                                                         & 10x10 & 5x5                                                   & 1      & ReLU                                                          \\ \hline
4      & \begin{tabular}[c]{@{}c@{}}Max Pooling\end{tabular} & 16                                                         & 5x5   & 2x2                                                   & 2      & -                                                             \\ \hline
5      & FC                                                    & -                                                          & 120   & -                                                     & -      & ReLU                                                          \\ \hline
6      & FC                                                    & -                                                          & 84    & -                                                     & -      & ReLU                                                          \\ \hline
Output & FC                                                    & -                                                          & 10    & -                                                     & -      & Softmax                                                       \\ \hline
\end{tabular}
\label{Lenet-architecture}
}

\end{table}

\subsection{VggNet}
We evaluate the effect of our proposed attacks on CIFAR-10 using the Vgg model. The detailed pre-trained Vgg network architecture is shown in Table~\ref{VGG-architecture}. It achieves a test accuracy of 81.05\% without poisoning. The model is trained with the Momentum optimizer with 0.9 momentum for 250 epochs. The learning rate starts at 0.01 and is scheduled with a decay rate of 0.5 every 25 epochs.

\begin{table}[htbp]
\caption{Network Architecture for the CIFAR10 experiments}
\scalebox{0.71}{
\begin{tabular}{|c|c|c|c|c|c|c|}
\hline
\multicolumn{2}{|c|}{Layer}                                    & \begin{tabular}[c]{@{}c@{}}Number\\of Filter\end{tabular} & Size    & \begin{tabular}[c]{@{}c@{}}Kernel\\ Size\end{tabular} & Stride & \begin{tabular}[c]{@{}c@{}}Activation\\ Function\end{tabular} \\ \hline
Input  & Image                                                 & -                                                          & 32x32x3 & -                                                     & -      & -                                                             \\ \hline
1      & Convolution                                           & 64                                                         & 32x32   & 3x3                                                   & 1      & ReLU                                                          \\ \hline
2      & Convolution                                           & 64                                                         & 32x32   & 3x3                                                   & 1      & ReLU                                                          \\ \hline
3      & \begin{tabular}[c]{@{}c@{}}Max Pooling\end{tabular} & 64                                                         & 16x16   & 2x2                                                   & 2      & -                                                             \\ \hline
4      & Convolution                                           & 128                                                        & 16x16   & 3x3                                                   & 1      & ReLU                                                          \\ \hline
5      & Covnvolution                                          & 128                                                        & 16x16   & 3x3                                                   & 1      & ReLU                                                          \\ \hline
6      & \begin{tabular}[c]{@{}c@{}}Max Pooling\end{tabular} & 128                                                        & 8x8     & 2x2                                                   & 2      & -                                                             \\ \hline
7      & Convolution                                           & 256                                                        & 8x8     & 3x3                                                   & 1      & ReLU                                                          \\ \hline
8      & Convolution                                           & 256                                                        & 8x8     & 3x3                                                   & 1      & ReLU                                                          \\ \hline
9      & \begin{tabular}[c]{@{}c@{}}Max Pooling\end{tabular} & 256                                                        & 4x4     & 2x2                                                   & 2      & -                                                             \\ \hline
10     & FC                                                    & -                                                          & 1024    & -                                                     & -      & ReLU                                                          \\ \hline
11     & FC                                                    & -                                                          & 180     & -                                                     & -      & ReLU                                                          \\ \hline
Output & FC                                                    & -                                                          & 10      & -                                                     & -      & Softmax                                                       \\ \hline
\end{tabular}
\label{VGG-architecture}
}
\end{table}
\subsection{ResNet}
For experiments on ImageNet, we use a pre-trained ResNet-50 model\footnote{https://github.com/keras-team/keras}. The benign model has a Top-1 accuracy of 74.87\% and Top-5 accuracy of 92.02\%, respectively. The training hyper-parameters and learning rate decay strategy follow the same settings as in the original paper [17].

\section{Additional Experiments of COEG Attack}
To have a comprehensive understanding of the effectiveness of COEG attack, we perform more in-depth experiments and analysis for both single instance attack and attack with a set of poisoned data.

\subsection{Single instance attack}
\subsubsection{Different class pairs.} Recent research [36] found that the effect of the poisoning integrity attacks (a.k.a backdoor) is highly dependent on class pairs and the choice of seed images. For instance, the same seed image with different poisoned labels (seed-target pairs) will result in a dramatically inconsistent poisoning effect. Meanwhile, different seed images with the same poisoned label will also lead to varied performances. Note that a different poisoned label means a different supplanter class in the COEG attack. To study whether the issue also exists in the availability poisoning attack, we perform two experiments for the single instance attack: i) we pick a different seed image from the previous experiment on MNIST and assign the same poisoned labels; ii) we pick the same seed image from CIFAR-10 and assign different poisoned labels. We keep the same training setting and baseline attacks as the previous experiments. The results of the first experiment are presented in Table~\ref{same-seed-different-label}. We find a similar trend that different labels yield different poisoning effect, although the difference is not as large as in poisoning integrity attacks [36]. As shown, our proposed attack still significantly outperform the baseline attack in test error and CTT rate.

\begin{table}[htbp]
\centering
\caption{Comparison of different seed images with the same poisoned label.}
\scalebox{1}{
\begin{tabular}{ccc}
\toprule
Attack (seed-label pairs)    & Test Error & CTT Rate \\ \hline
FL attack (seed:1-label:4)   & 27.59\%    & 27.33\%  \\
\textbf{Proposed attack (1-4)} & 71.11\%    & 79.90\%  \\ \midrule
FL attack (seed:6-label:4)   & 19.00\%    & 32.85\%  \\
\textbf{Proposed attack (6-4)} & 85.50\%    & 85.85\%  \\ \bottomrule
\label{same-seed-different-label}
\end{tabular}}
\end{table}



\begin{figure}[htbp]
    \centering
    \resizebox{0.48\textwidth}{!}{
    \includegraphics{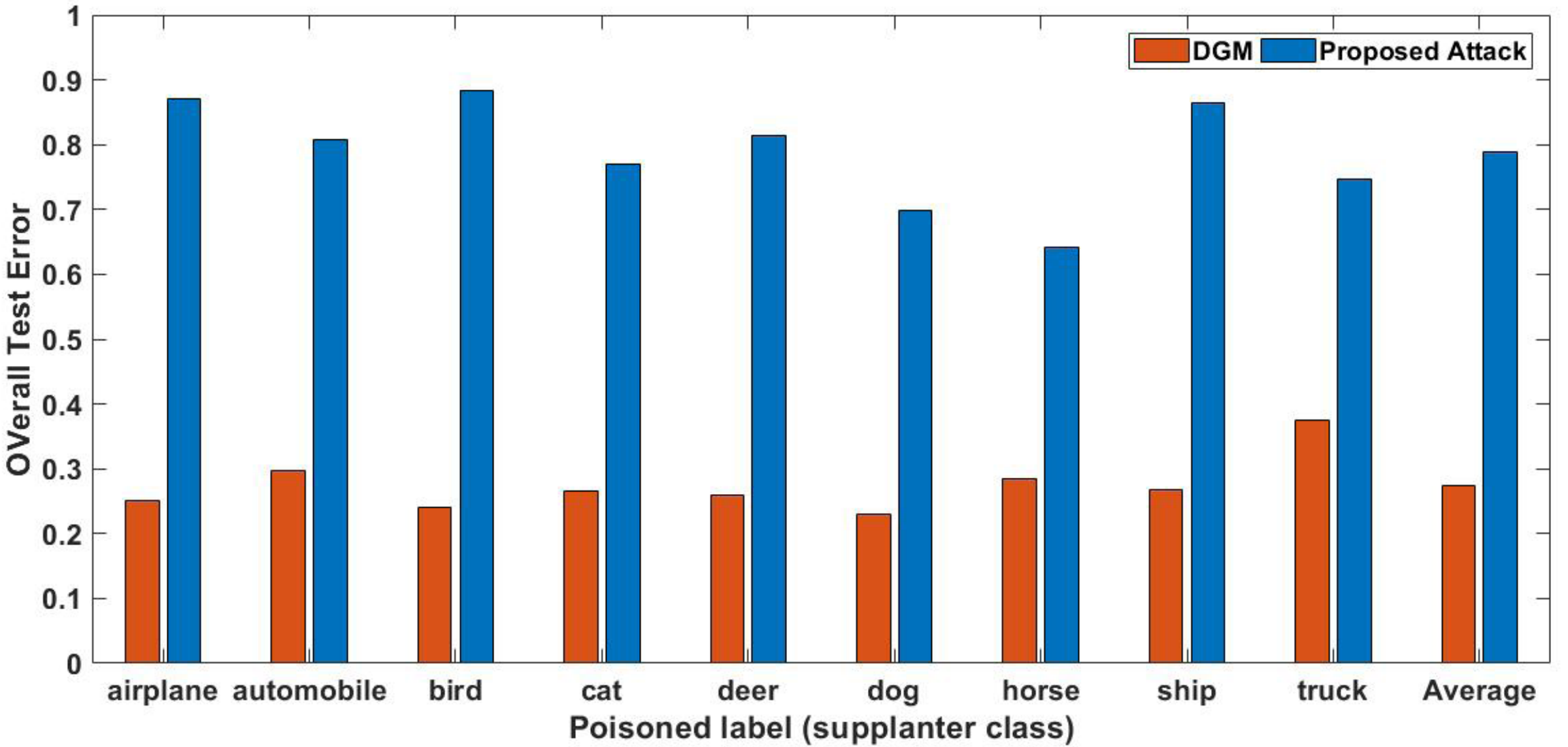}}
    \caption{Test error with different seed-target pairs on CIFAR-10 for the single instance attack.}
    \label{error-bar-cifar10-target-pairs}
\end{figure}

\begin{figure}[htbp]
    \centering
    \resizebox{0.48\textwidth}{!}{
    \includegraphics{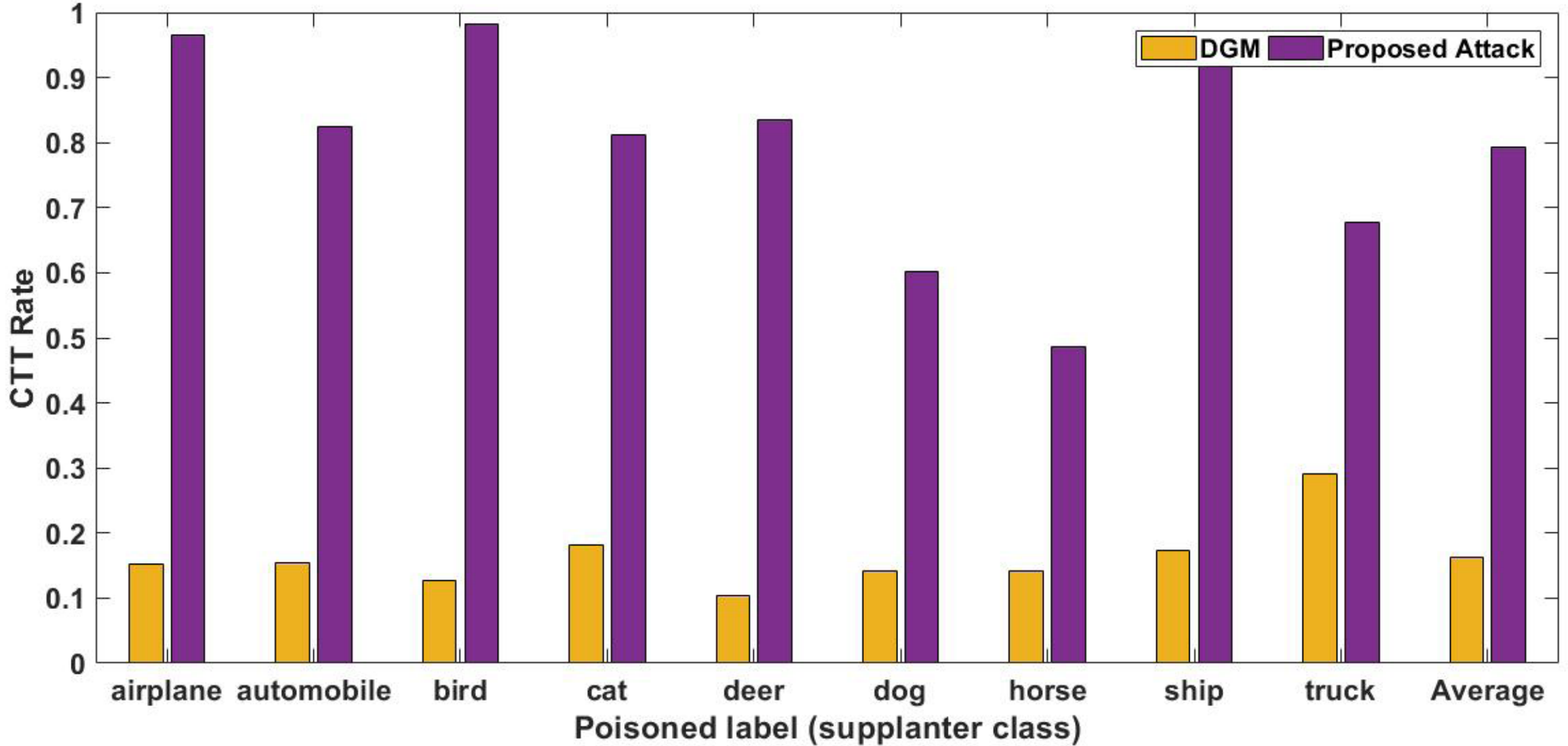}}
    \caption{CTT rate with different seed-target pairs on CIFAR-10 for the single instance attack.}
    \label{ctt-bar-cifar10-target-pairs}
\end{figure}

The results of the second experiment are presented in Figure~\ref{error-bar-cifar10-target-pairs} and Figure~\ref{ctt-bar-cifar10-target-pairs}. Although some of the class pairs perform a bit worse than others (i.e., frog-cat for the baseline attack, frog-horse for our proposed attack), both the FL attack and our proposed attack show a relatively consistent performance with different poisoned labels. The proposed COEG attack achieves an test error of 78.87\% and CTT rate of 79.20\% on average, which significantly outperforms the baseline attack 
with an average test error of 27.44\% and CTT rate of 16.30\%. Note that the CTT of our approach is 65.9\% higher than that of the baseline attack, which exceeds the performance difference of test error. This fact further validates that our proposed attack is particularly effective as class-oriented poisoning attacks.

\subsubsection{Attack with more images.} While lower learning rates can reduce the effect of poisoning attacks, the limited number of poisoned data may also be a factor. To better understand the limitation of the single instance attack, we keep the lower learning rate and increase the number of poisoned data by duplicating the single poisoned sample. The results are summarized in Table~\ref{attack-with-more-images}. With the increase of poisoned data, the overall test error and CTT rate of the baseline attack remain nearly the same while ours even slightly drop by 13\% and 7\%, respectively. These results reveals that simply increasing the number of poisoned data by duplication will not improve the poisoning effect. Thus, the diversity of the poisoned dataset is crucial. 

\begin{table}[htbp]
\caption{Comparison of poisoning effect with different numbers of poisoned data.}
\begin{tabular}{c|cccc}
\toprule
\multirow{2}{*}{\begin{tabular}[c]{@{}c@{}}Number of\\ poisoned data\end{tabular}} & \multicolumn{2}{c}{Test Error} & \multicolumn{2}{c}{CTT Rate} \\ \cline{2-5} 
                                                                                   & DGM            & Ours          & DGM           & Ours         \\ \hline
1                                                                                  & 23.4\%         & 54.8\%        & 11.52\%       & 23.32\%      \\
100                                                                                & 22.9\%         & 41.1\%        & 10.7\%        & 16.8\%       \\
500                                                                                & 22.6\%         & 41.1\%        & 10.6\%        & 16.5\%       \\ \bottomrule
\end{tabular}
\label{attack-with-more-images}
\end{table}

\subsection{Attack with a set of poisoned data}
\subsubsection{Different target supplanter classes.} 
\begin{figure}[htbp]
    \centering
    \resizebox{0.48\textwidth}{!}{
    \includegraphics{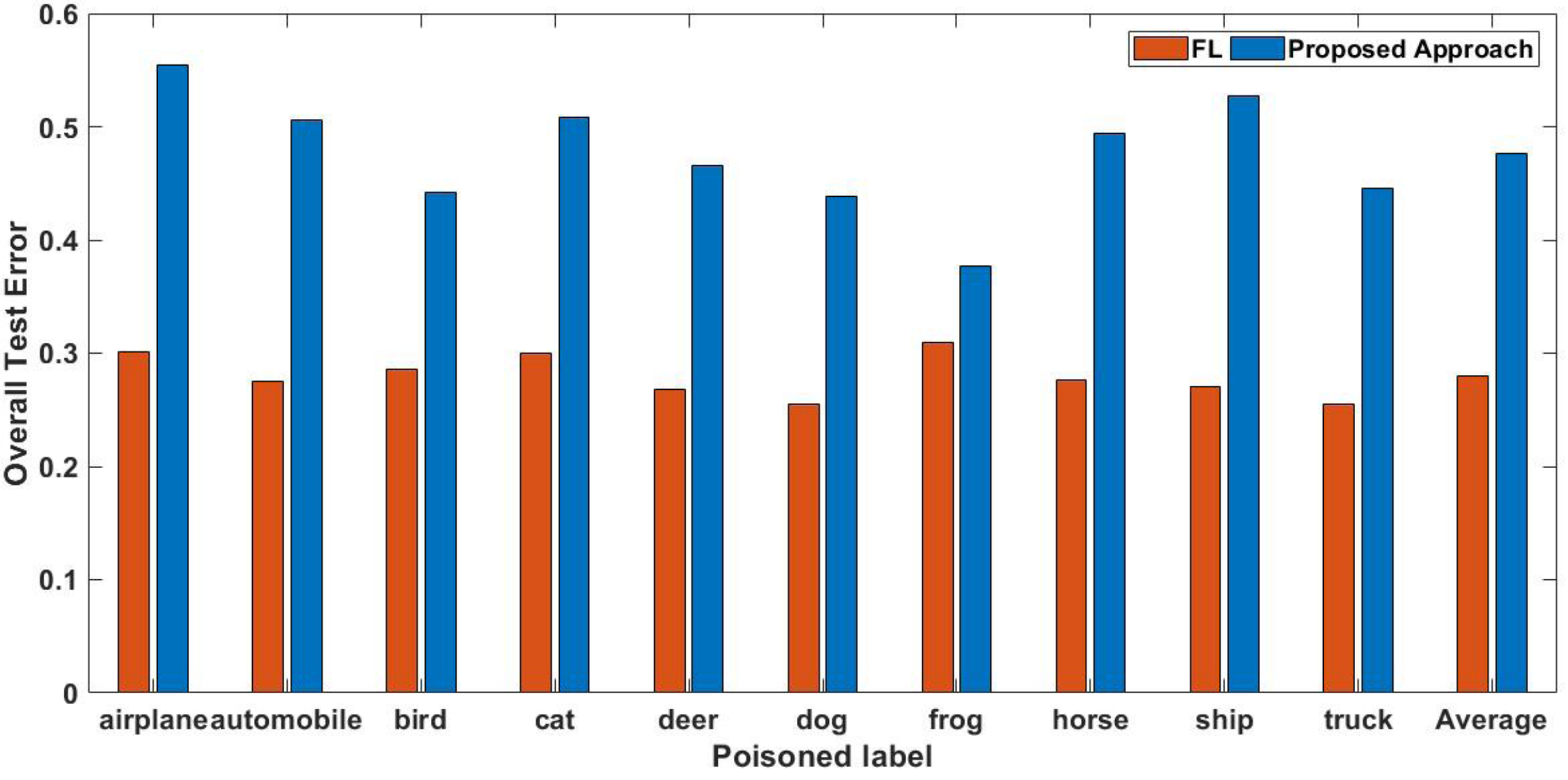}}
    \caption{Test error with different target labels on CIFAR10 for the attack with a set of poisoned data.}
    \label{error-bar-cifar10-target-pairs-general}
\end{figure}

\begin{figure}[htbp]
    \centering
    \resizebox{0.48\textwidth}{!}{
    \includegraphics{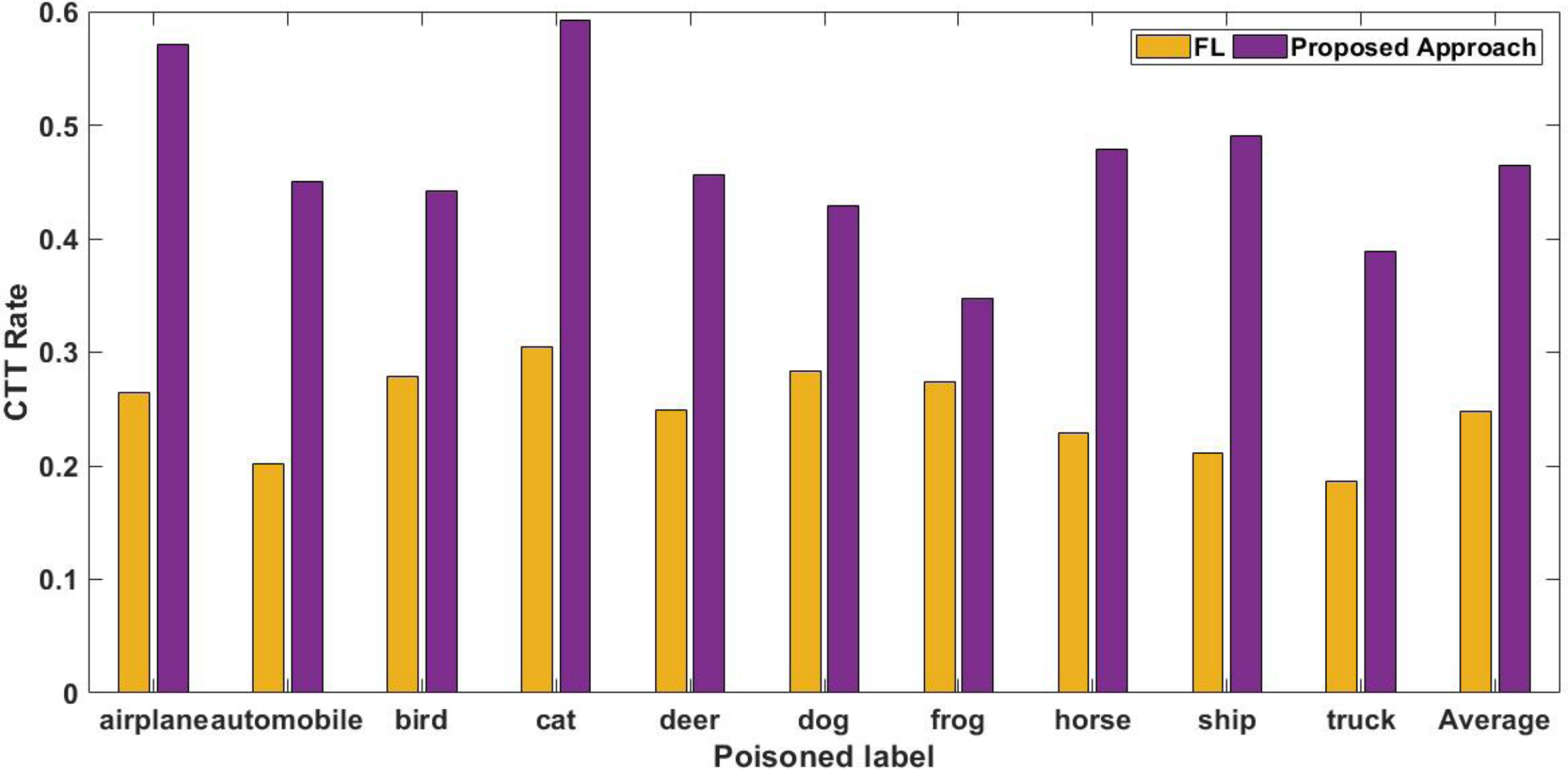}}
    \caption{CTT rate with different target labels on CIFAR10 for the attack with a set of poisoned data.}
    \label{ctt-bar-cifar10-target-pairs-general}
\end{figure}

We study the impact of different target supplanter classes for the general poisoning attack, where a set of poisoned data is injected. This setting is slightly different from the single instance attack as the poisoned seed images are arbitrarily selected from all classes. We assign different targeted labels (supplanter classes) to the poisoned data and compare the performance between the FL attack and our proposed attack. The test error and CTT rate are presented in Figure~\ref{error-bar-cifar10-target-pairs-general} and Figure~\ref{ctt-bar-cifar10-target-pairs-general}. We use $\alpha = 0.5$ and a learning rate of $1 \times 10^{-5}$. Similar to the single instance attack, our proposed attack is resilient to the variation of poisoned labels and achieves the class-oriented adversarial goal. 

\subsubsection{Different dataset sizes.} Another interesting finding in paper [36] is that the effect of poisoning attacks is related to the size of the retraining dataset. For instance, with a fixed percentage of poisoned data, some attacks achieve better poisoning effect on a larger retraining dataset, while others perform worse. The impact of dataset size is particularly worth studying in the scenario of end-to-end fine-tuning on a pre-trained model, where retraining data are scarce. We test three different sizes with the same percentage of poisoned data for each size. We use the same training setting as above. The results are presented in Tables~\ref{error-different-dataset-size} and~\ref{ctt-different-dataset-size}. 

It can be seen that a larger dataset size always yields better poisoning performance at all percentage levels for the baseline FL attack. However, our proposed attack has the same trend when the percentage of poisoned data is small and shows superior performance on the dataset size of 500. When the dataset size is small, the FL attack is barely effective with any percentage of poisoned data, while the proposed attack achieves improved test error and CTT rate. Moreover, the performance improvement of FL is much smaller than the proposed approach with the increase of dataset size. These observations provide sufficient evidence that our proposed attack is much more effective than the baseline attack and more elastic to dataset size change.

\begin{table}[htbp]
\centering
\caption{Comparison of test error with different dataset sizes (lr = $1 \times 10^{-5}$).}
\label{error-different-dataset-size}
\scalebox{0.76}{
\begin{tabular}{c|cc|cc|cc}
\toprule
\begin{tabular}[c]{@{}c@{}}Test\\ Error\end{tabular} & \multicolumn{2}{c|}{\begin{tabular}[c]{@{}c@{}}Training dataset \\ size = 100\end{tabular}} & \multicolumn{2}{c|}{\begin{tabular}[c]{@{}c@{}}Training dataset\\ size = 500\end{tabular}} & \multicolumn{2}{c}{\begin{tabular}[c]{@{}c@{}}Training dataset\\ size = 1000\end{tabular}} \\ \hline
                                                     & FL                                          & \textbf{Ours}                                          & FL                                          & \textbf{Ours}                                         & FL                                           & \textbf{Ours}                                        \\
$\alpha = 0.1$                                              & 18.95\%                                            & \textbf{18.97\%}                                               &19.21\%                                             &\textbf{23.84\%}                                              & 19.65\%                                      & \textcolor{red}{\textbf{36.38\%} }                                    \\
$\alpha = 0.2$                                               & 18.92\%                                            &\textbf{19.29\%}                                               &19.54\%                                             &\textbf{37.74\%}                                              & 20.84\%                                      & \textcolor{red}{\textbf{47.00\%}}                                     \\
$\alpha = 0.3$                                               & 18.79\%                                             &\textbf{19.64\%}                                               &20.37\%                                             &\textcolor{red}{\textbf{54.49\%}}                                              & 23.05\%                                      & \textbf{49.53\%}                                     \\
$\alpha = 0.4$                                               & 18.92\%                                            & \textbf{20.70\%}                                              &21.20\%                                             &\textcolor{red}{\textbf{64.23\%}}                                              & 26.10\%                                      & \textbf{51.84\%}                                     \\
$\alpha = 0.5$                                               &18.99\%                                             &\textbf{22.27\%}                                               &22.41\%                                             &\textcolor{red}{\textbf{70.15\%}}                                              & 30.14\%                                        & \textbf{55.42\%}\\     \bottomrule                               
\end{tabular}
}
\end{table}

\begin{table}[htbp]
\centering
\caption{Comparison of CTT rate with different dataset sizes (lr = $1 \times 10^{-5}$).}
\label{ctt-different-dataset-size}
\scalebox{0.76}{
\begin{tabular}{c|cc|cc|cc}
\toprule
\begin{tabular}[c]{@{}c@{}}CTT\\ Rate\end{tabular} & \multicolumn{2}{c|}{\begin{tabular}[c]{@{}c@{}}Training dataset \\ size = 100\end{tabular}} & \multicolumn{2}{c|}{\begin{tabular}[c]{@{}c@{}}Training dataset\\ size = 500\end{tabular}} & \multicolumn{2}{c}{\begin{tabular}[c]{@{}c@{}}Training dataset\\ size = 1000\end{tabular}} \\ \hline
                                                   & FL                                      & \textbf{Ours}                                     & FL                                     & \textbf{Ours}                                     & FL                                      & \textbf{Ours}                                    \\
$\alpha = 0.1$                                            & 10.35\%                                        & \textbf{10.76\%}                                                &11.22\%                                        & \textbf{16.00\%}                                                  & 12.11\%                                 & \textcolor{red}{\textbf{31.25\%}}                                 \\
$\alpha = 0.2$                                            &10.50\%                                         &\textbf{11.46\%}                                                  &12.20\%                                        &\textbf{33.46\%}                                                   & 14.46\%                                 & \textcolor{red}{\textbf{46.70\%}}                                 \\
$\alpha = 0.3$                                            &10.06\%                                         &\textbf{12.25\%}                                                   &13.64\%                                        &\textcolor{red}{\textbf{56.83\%}}                                                   & 17.48\%                                 & \textbf{49.75\%}                                 \\
$\alpha = 0.4$                                            & 10.80\%                                        &\textbf{13.28\%}                                                   &14.83\%                                        &\textcolor{red}{\textbf{69.46\%}}                                                   & 21.4\%                                  & \textbf{52.43\%}                                 \\
$\alpha = 0.5$                                            &10.98\%                                         &\textbf{14.88\%}                                                   &16.57\%                                        &\textcolor{red}{\textbf{76.75\%}}                                                   & 26.45\%                                 & \textbf{57.11\%}\\\bottomrule                                
\end{tabular}
}
\end{table}

\begin{table*}[hbtp]
\centering
\caption{CFT rate of each CIFAR-10 class by poisoning with different $\alpha$ at learning rate $1 \times 10^{-5}$.}
\label{CFT-CIFAR10-different-alpha}
\scalebox{0.9}{
\begin{tabular}{cccccccccccc}
\toprule
\begin{tabular}[c]{@{}c@{}}Fraction of \\ poisoned data\end{tabular} & \begin{tabular}[c]{@{}c@{}}Attacks
\end{tabular} & airplane & automobile & bird    & cat     & deer    & dog     & frog    & horse   & ship    & \begin{tabular}[c]{@{}c@{}}\cellcolor[HTML]{C0C0C0}{truck}\\ \cellcolor[HTML]{C0C0C0}{(victim)}\end{tabular} \\ \midrule
\multirow{3}{*}{$\alpha = 0.1$}                                         & FL-1                                                    & -5.57\%  & -0.99\%    & -0.27\% & -0.65\% & 5.53\%  & 2.21\%  & 3.53\%  & -0.69\% & 1.23\%  & \cellcolor[HTML]{C0C0C0}{2.75\%}                                                   \\
                                                                     & FL-2                                                    & -1.62\%  & -0.99\%    & -0.68\% & -1.47\% & 2.19\%  & 0.55\%  & 0.35\%  & -0.69\% & -0.78\% & \cellcolor[HTML]{C0C0C0}{1.83\%}                                                   \\
                                                                     & \textbf{Ours}                                           & -3.48\%  & -1.99\%    & -0.82\% & -3.11\% & 3.47\%  & 1.11\%  & -0.35\% & -0.23\% & -0.67\% & \cellcolor[HTML]{C0C0C0}{\textcolor{red}{7.89\%}}                                                   \\ \midrule
\multirow{3}{*}{$\alpha = 0.2$}                                         & FL-1                                                    & -9.27\%  & -0.44\%    & 2.73\%  & 0.49\%  & 11.05\% & 4.01\%  & 6.35\%  & -0.34\% & 4.37\%  & \cellcolor[HTML]{C0C0C0}{3.66\%}                                                   \\
                                                                     & FL-2                                                    & -2.20\%  & -0.88\%    & -1.50\% & -1.47\% & 1.67\%  & 1.24\%  & -0.12\% & -0.69\% & -1.23\% & \cellcolor[HTML]{C0C0C0}{3.20\%}                                                   \\
                                                                     & \textbf{Ours}                                           & -4.17\%  & -1.10\%    & -0.82\% & -3.11\% & 3.37\%  & 2.21\%  & -0.12\% & 0.23\%  & -0.56\% & \cellcolor[HTML]{C0C0C0}{\textcolor{red}{15.79\%}}                                                  \\ \midrule
\multirow{3}{*}{$\alpha = 0.3$}                                         & FL-1                                                    & -11.24\% & 0.11\%     & 5.05\%  & 3.27\%  & 17.10\% & 7.88\%  & 9.75\%  & 1.49\%  & 7.05\%  & \cellcolor[HTML]{C0C0C0}{8.01\%}                                                   \\
                                                                     & FL-2                                                    & -2.67\%  & -1.10\%    & -1.91\% & -1.47\% & 2.57\%  & 1.66\%  & -0.24\% & -0.57\% & -1.01\% & \cellcolor[HTML]{C0C0C0}{5.03\%}                                                   \\
                                                                     & \textbf{Ours}                                           & -5.01\%  & -1.66\%    & -1.23\% & -2.13\% & 5.91\%  & 3.04\%  & 0.24\%  & 0.91\%  & -0.67\% & \cellcolor[HTML]{C0C0C0}{\textcolor{red}{28.60\%}}                                                  \\ \midrule
\multirow{3}{*}{$\alpha = 0.4$}                                         & FL-1                                                    & -12.98\% & 0.55\%     & 10.52\% & 6.71\%  & 24.29\% & 11.20\% & 17.16\% & 4.46\%  & 11.09\% & \cellcolor[HTML]{C0C0C0}{12.01\%}                                                  \\
                                                                     & FL-2                                                    & -3.13\%  & -1.10\%    & -1.50\% & -2.29\% & 2.44\%  & 2.77\%  & 0.00\%  & -0.67\% & -1.01\% & \cellcolor[HTML]{C0C0C0}{5.84\%}                                                   \\
                                                                     & \textbf{Ours}                                           & -6.37\%  & -0.33\%    & -0.27\% & -0.49\% & 6.43\%  & 3.32\%  & 0.47\%  & 0.69\%  & -0.67\% & \cellcolor[HTML]{C0C0C0}{\textcolor{red}{38.10\%}}                                                  \\ \midrule
\multirow{3}{*}{$\alpha = 0.5$}                                         & FL-1                                                    & -13.09\%  & 3.87\%     & 16.80\% & 16.20\% & 32.65\% & 19.09\% & 22.44\% & 8.34\%  & 15.45\% & \cellcolor[HTML]{C0C0C0}{18.42\%}                                                  \\
                                                                     & FL-2                                                    & -3.01\%  & -1.44\%    & -2.05\% & -2.29\% & 2.57\%  & 2.67\%  & -0.59\% & -0.11\% & -0.78\% & \cellcolor[HTML]{C0C0C0}{8.01\%}                                                   \\
                                                                     & \textbf{Ours}                                           & -7.07\%  & 0.88\%     & -0.41\% & 2.29\%  & 5.27\%  & 3.32\%  & -0.47\% & 0.11\%  & 0.22\%  & \cellcolor[HTML]{C0C0C0}{\textcolor{red}{51.14\%}}                                                 \\ \bottomrule
\end{tabular}}
\end{table*}

\section{Additional Experiments of COES Attack}

As discussed in the main manuscript, the adversarial goal for the COES attack is more challenging than COEG. We also extensively study the effectiveness of our proposed approach on the COES attack with more experiments.

\subsection{Different values of $\boldsymbol{\alpha}$.} We first study the impact of various $\alpha$ values on the COES attack. We keep the same training setting as in previous experiments. The results are presented in Table~\ref{CFT-CIFAR10-different-alpha}. As expected, we have a similar finding as the COEG attack that the poisoning effect is proportional to the number of injected poisoned data. On the other hand, we find that for all different $\alpha$, FL-1 always performs badly in reducing the CFT rates of non-victim classes, while FL-2 reduces the CFT rates of non-victim classes at the cost of lowering the CFT rate of the victim class at the same time. Our proposed approach addresses for the shortcomings of both FL-1 and FL-2, achieving a high CFT rate for the victim class while keeping low CFT rates for non-victim classes.

\subsection{Different learning rates.}
We present the results of different learning rates of the COES attack in Table~\ref{CFT-CIFAR10-different-lr}. We keep the dataset size at 1000 and set the $\alpha$ value to 0.5. Our proposed approach consistently outperforms the baseline attacks. With the increase of the learning rate, the CFT rates of the victim class increase for all three attacks. Interestingly, our proposed attack achieves the best CFT rate of the victim class at a learning rate of $4 \times 10^{-5}$ other than a higher learning rate. In comparison, with the decrease of the learning rate, FL-1 and FL-2 suffer drastic performance drops. Thus, it can be concluded that the proposed approach is more resilient to the variation of the learning rate for the COES attack.

\subsection{Different training dataset sizes.} Lastly, we evaluate the impact of different training dataset sizes. We keep the learning rate at $1 \times 10^{-5}$ and $\alpha$ value at 0.5 for the experiment and present the results in Table~\ref{CFT-CIFAR0-different-size}. Unlike the COEG attack that achieves a better CTT performance with a dataset size of 500, a larger dataset size in COES attack always yields a better CFT rate of the victim class without significantly affecting the CFT rates of non-victim classes. Note that the poisoning effect is greatly reduced by scaling down the training dataset size. This further shows the challenging adversarial goal of the COES attack as we need to manipulate the performance of all classes with limited poisoned data. However, even training on an extremely small dataset, we can still achieve nearly 5 times better performance than the FL-2 attack, which is a clear showcase of the effectiveness of our proposed attacks.


\begin{table*}[ht!]
\caption{CFT rate of each CIFAR-10 class by poisoning with different learning rates when $\alpha = 0.5$.}
\label{CFT-CIFAR10-different-lr}
\scalebox{0.93}{
\begin{tabular}{cccccccccccc}
\toprule
\begin{tabular}[c]{@{}c@{}}Learning\\ Rate\end{tabular}             & Attacks       & airplane & automobile & bird    & cat     & deer    & dog     & frog    & horse   & ship    & \begin{tabular}[c]{@{}c@{}}\cellcolor[HTML]{C0C0C0}{truck}\\ \cellcolor[HTML]{C0C0C0}{(victim)}\end{tabular} \\ \midrule
\multirow{3}{*}{$1 \times 10^{-4}$} & FL-1          & -14.60\% & 26.85\%    & 45.49\% & 37.15\% & 48.97\% & 46.47\% & 46.65\% & 36.80\% & 43.23\% & \cellcolor[HTML]{C0C0C0}{53.66\%}                                                  \\
                                                                    & FL-2          & -6.84\%  & 0.55\%     & 0.82\%  & -6.55\% & 3.21\%  & -0.55\% & -2.82\% & -0.34\% & 0.90\%  & \cellcolor[HTML]{C0C0C0}{41.30\%}                                                  \\
                                                                    & \textbf{Ours} & -1.85\%  & -1.22\%    & 3.28\%  & -6.71\% & 2.70\%  & -0.69\% & -1.06\% & 2.40\%  & -2.35\% & \cellcolor[HTML]{C0C0C0}{\textcolor{red}{65.90\% }}                                                 \\ \midrule
\multirow{3}{*}{$4 \times 10^{-5}$} & FL-1          & -14.83\% & 22.54\%    & 39.21\% & 39.28\% & 54.24\% & 39.42\% & 49.24\% & 31.66\% & 40.76\% & \cellcolor[HTML]{C0C0C0}{48.86\%}                                                  \\
                                                                    & FL-2          & -6.72\%  & 0.33\%     & -1.23\% & -3.93\% & 4.24\%  & 3.04\%  & -0.65\% & -0.34\% & 1.34\%  & \cellcolor[HTML]{C0C0C0}{25.51\%}                                                  \\
                                                                    & \textbf{Ours} & -3.01\%  & 0.88\%     & 1.91\%  & -2.45\% & 3.47\%  & -0.28\% & 0.24\%  & 2.29\%  & -2.02\% & \cellcolor[HTML]{C0C0C0}{\textcolor{red}{66.59\% }}                                                 \\ \midrule
\multirow{3}{*}{$5 \times 10^{-6}$} & FL-1          & -10.66\% & -0.22\%    & 4.37\%  & 1.31\%  & 15.55\% & 7.47\%  & 8.70\%  & 1.37\%  & 5.38\%  & \cellcolor[HTML]{C0C0C0}{7.21\% }                                                  \\
                                                                    & FL-2          & -1.74\%  & -0.88\%    & -2.05\% & -1.31\% & 1.67\%  & 1.52\%  & 0.12\%  & -0.46\% & -0.78\% & \cellcolor[HTML]{C0C0C0}{4.46\% }                                                  \\
                                                                    & \textbf{Ours} & -5.10\%  & -1.10\%    & -2.19\% & -0.16\% & 3.21\%  & 3.04\%  & 0.12\%  & 0.80\%  & -0.90\% & \cellcolor[HTML]{C0C0C0}{\textcolor{red}{27.80\%}}                                                  \\ \bottomrule
\end{tabular}}
\end{table*}

\begin{table*}[ht!]
\caption{CFT rate of each CIFAR-10 class by poisoning with different training dataset sizes.}
\label{CFT-CIFAR0-different-size}
\scalebox{0.95}{
\begin{tabular}{cccccccccccc}
\toprule
\begin{tabular}[c]{@{}c@{}}Dataset\\ size\end{tabular} & Attacks       & airplane & automobile & bird    & cat     & deer    & dog     & frog    & horse   & ship    & \begin{tabular}[c]{@{}c@{}}\cellcolor[HTML]{C0C0C0}{truck}\\ \cellcolor[HTML]{C0C0C0}{(victim)}\end{tabular} \\ \midrule
\multirow{3}{*}{100}                                             & FL-1          & -2.90\%  & -0.55\%    & -1.91\% & -1.96\% & 2.96\%  & 1.80\% & 1.76\%  & -0.69\% & -0.67\% & \cellcolor[HTML]{C0C0C0}{1.83\%}                                                   \\
                                                                & FL-2          & -0.93\%  & -0.33\%    & -0.68\% & -0.16\% & 0.51\%  & 0.55\% & 0.47\%  & -0.34\% & -0.34\% & \cellcolor[HTML]{C0C0C0}{1.14\%}                                                   \\
                                                                & \textbf{Ours} & -1.97\%  & 0.00\%     & -1.23\% & -1.64\% & 1.41\%  & 1.94\% & -0.24\% & -0.11\% & -0.67\% & \cellcolor[HTML]{C0C0C0}{\textcolor{red}{4.92\%} }                                                  \\ \midrule
\multirow{3}{*}{500}                                            & FL-1          & -10.66\% & -0.33\%    & 3.69\%  & 1.96\%  & 14.65\% & 8.30\%  & 8.81\%  & 1.03\%  & 5.28\%  & \cellcolor[HTML]{C0C0C0}{7.09\%  }                                                 \\
                                                                & FL-2          & -2.67\%  & -0.66\%    & -2.19\% & -1.15\% & 1.54\%  & 3.18\%  & 0.59\%  & -0.69\% & -0.78\% & \cellcolor[HTML]{C0C0C0}{4.35\%}                                                   \\
                                                                & \textbf{Ours} & -4.75\%  & -0.88\%    & -1.50\% & -3.11\% & 3.73\%  & 3.32\%  & 0.94\%  & 0.69\%  & -0.67\% & \cellcolor[HTML]{C0C0C0}{\textcolor{red}{24.03\%}}                                                \\ \midrule
\multirow{3}{*}{1000}                                           & FL-1                                                    & -13.09\%  & 3.87\%     & 16.80\% & 16.20\% & 32.65\% & 19.09\% & 22.44\% & 8.34\%  & 15.45\% & \cellcolor[HTML]{C0C0C0}{18.42\%}                                                  \\
                                                                     & FL-2                                                    & -3.01\%  & -1.44\%    & -2.05\% & -2.29\% & 2.57\%  & 2.67\%  & -0.59\% & -0.11\% & -0.78\% & \cellcolor[HTML]{C0C0C0}{8.01\%}                                                   \\
                                                                     & \textbf{Ours}                                           & -7.07\%  & 0.88\%     & -0.41\% & 2.29\%  & 5.27\%  & 3.32\%  & -0.47\% & 0.11\%  & 0.22\%  & \cellcolor[HTML]{C0C0C0}{\textcolor{red}{51.14\%}}                                                 \\ \bottomrule
\end{tabular}}
\end{table*}



\end{document}